# Long Story Short: Story-level Video Understanding from 20K Short Films

**Ridouane Ghermi[1] · Xi Wang[1] · Vicky Kalogeiton[1] · Ivan Laptev[2]**

## Abstract

Recent developments in vision-language models have significantly advanced video understanding. Existing datasets and tasks, however, have notable limitations. Most datasets are confined to short videos with limited events and narrow narratives. For example, datasets with instructional and egocentric videos often depict the activities of one person in a single scene. Although existing movie datasets offer richer content, they are often limited to short-term tasks, lack publicly available videos, and frequently encounter *data leakage* issues given the use of subtitles and other information about commercial movies during LLM pretraining. To address the above limitations, we propose **Short-Films 20K (SF20K)**, the largest publicly available movie dataset. SF20K consists of 20,143 amateur films, amounting to 3,582 hours of video, with an average of 12 minutes per movie. We accompany this dataset with **SF20K-Test**, a manual, open-ended question answering benchmark. SF20K-Test consists of 95 movies and 979 question-answer pairs. Our extensive analysis of SF20K-Test reveals limited data leakage, emphasizes the need for long-term reasoning, and demonstrates the strong performance of recent VLMs. Finally, we show that instruction tuning on the large-scale dataset substantially improves model performance, paving the way for future progress in long-term video understanding.

## 1 Introduction

Recent advances in vision-language models (Lin et al., 2023; Li et al., 2023; Chen et al., 2023a, b; Wang et al., 2022; Fu et al., 2023, 2022; Dai et al., 2023; Xu et al., 2023) show significant promise for enhancing machine perception. Despite their remarkable progress, current datasets are still constrained by some limitations. Most of them consist of short videos (Gu et al., 2018; Abu-El-Haija et al., 2016; Diba et al., 2020; Xu et al., 2016; Yang et al., 2022; Goyal et al., 2017; Soomro et al., 2012; Kay et al., 2017; Li et al., 2020; Grunde-Mclaughlin et al., 2021; Xiao et al., 2021), typically under one minute, and primarily target short-term tasks such as action recognition and video retrieval, which often require only a few seconds of visual content to solve (Mangalam et al., 2023; Lei et al., 2022). Long-form video understanding, on the other hand, has been mainly studied in three domains: (i) *Egocentric videos*, featuring extended first-person sequences with continuous actions (Grauman et al., 2022a; Damen et al., 2018; Jia et al., 2022; Lin et al., 2022; Mangalam et al., 2023; Sigurdsson et al., 2018; Sharghi et al., 2017) and daily activities ranging from object manipulation to social interactions (Grauman et al., 2022a). Although these datasets excel in video duration, they lack narrative depth, focusing on immediate visuals rather than rich storytelling; (ii) *Instructional videos* (Miech et al., 2019a; Yang et al., 2021), which depict a variety of human tasks and focus on the procedural aspect, are often brief and deficient in storytelling; (iii) *Movies*, unlike egocentric and instructional videos, provide complex plots and extended durations, making them

✉ Ridouane Ghermi
ridouane.ghermi@inria.fr

Xi Wang
xi.wang@polytechnique.edu

Vicky Kalogeiton
vicky.kalogeiton@polytechnique.edu

Ivan Laptev
ivan.laptev@inria.fr

1 LIX, Ecole Polytechnique, IP Paris, Palaiseau, France

2 MBZUAI, Abu Dhabi, United Arab Emirates

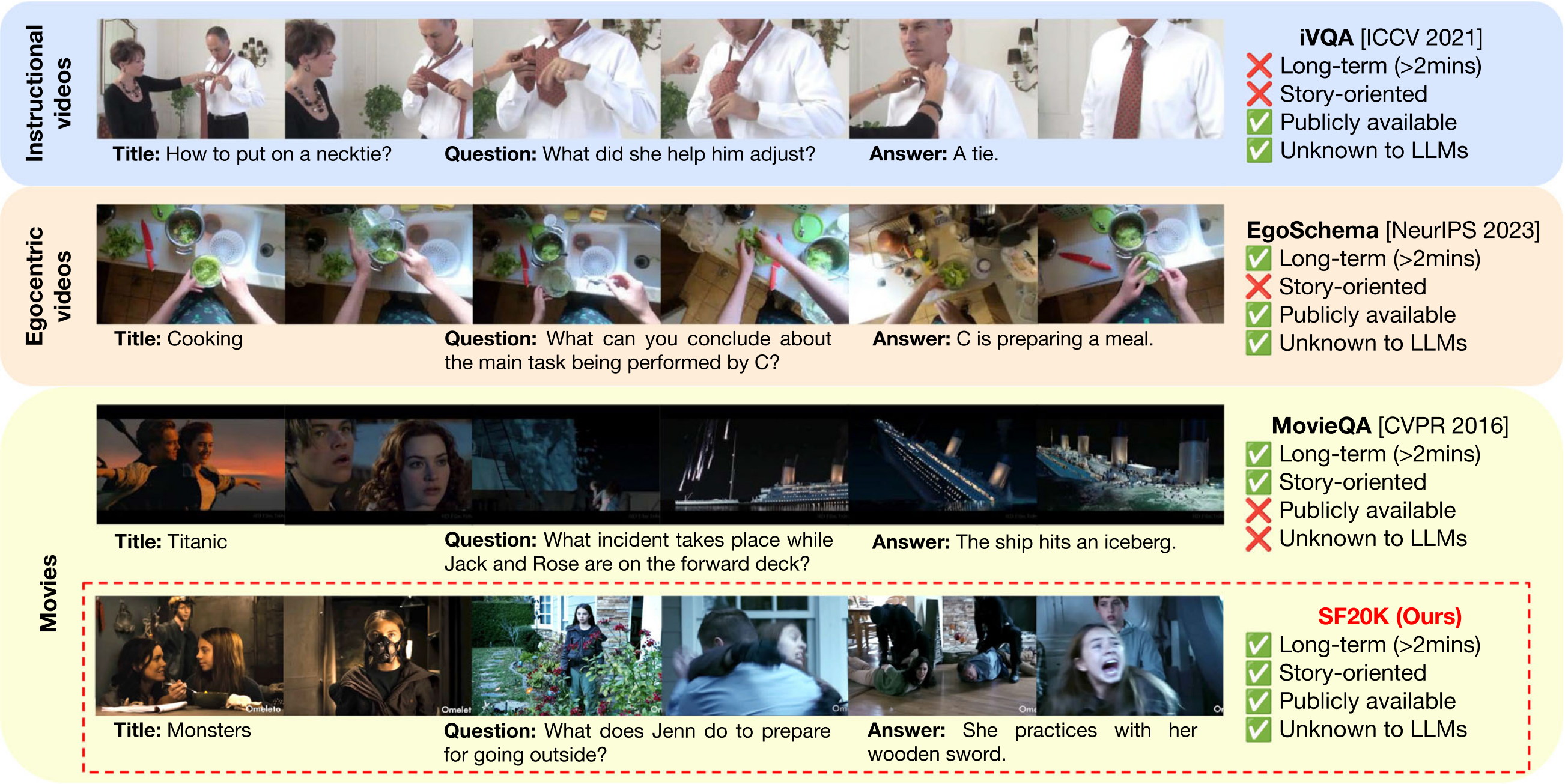

**Fig. 1** Examples of Video Question Answering (VideoQA) tasks across three domains: instructional videos, egocentric videos, and movies. While instructional and egocentric videos usually depict one or two people performing a single task, movies present time-extended stories with rich variety in terms of scenes, characters, and interactions

well suited for developing models capable of comprehending long-form narratives (see Figure 1).

Several movie datasets have been proposed in the literature (Rohrbach et al., 2016; Bain et al., 2020a; Tapaswi et al., 2016; Lei et al., 2019; Soldan et al., 2022; Han et al., 2023; Wu & Krähenbühl, 2021; Huang et al., 2020a; Vicol et al., 2018; Xiong et al., 2019; Sadhu et al., 2021; Marszałek et al., 2009; Rao et al., 2020; Chen et al., 2022; Bose et al., 2022; Maharaj et al., 2017; Song et al., 2023; Rawal et al., 2024b; Xie et al., 2024) offering rich, narrative-driven content and supporting various tasks. These tasks include clip retrieval (Bain et al., 2020a), video question answering (Tapaswi et al., 2016; Lei et al., 2019; Maharaj et al., 2017; Song et al., 2023; Rawal et al., 2024b), audio description (Soldan et al., 2022; Han et al., 2023; Xie et al., 2024), semantic role labeling (Sadhu et al., 2021), scene segmentation (Huang et al., 2020a; Rao et al., 2020), and visual scene recognition (Bose et al., 2022). However, current movie datasets suffer from three main limitations: (i) *Accessibility*: access to commercial movies is subject to copyright restrictions ; (ii) *Clip duration*: they often consist of incomplete and short clips (Tapaswi et al., 2016; Lei et al., 2019; Rawal et al., 2024a; Wu & Krähenbühl, 2021), often less than four minutes in length , limiting their potential for long-form understanding of evolving storylines. Moreover, these datasets frequently suffer from ill-designed benchmark questions with insufficient narrative focus (Song et al., 2023); (iii) *Data leakage*: as these datasets feature well-known commercial movies, modern Large Language Models (LLMs) and Vision-Language Models (VLMs) have likely been exposed to related content, such as synopses, reviews, discussions, subtitles, and blog posts. Figure 2 showcases this issue, showing that modern LLMs can achieve high accuracy on questions from major movie datasets using only movie titles. This leakage leads to biased benchmarks and ineffective training.

To address the limitations of existing movie datasets, we introduce **Short-Films 20K (SF20K)**, a novel video dataset comprising 20,143 short films totaling almost 3,582 hours of video, with an average duration of 12 minutes per film – significantly longer than existing datasets . Unlike copyrighted films, SF20K consists of publicly accessible amateur films across diverse genres sourced from YouTube and Vimeo. While shorter than traditional movies, these films feature complex narratives that unfold through key sequences of events and multiple character interactions. Crucially, with minimal exposure to LLMs, SF20K is less prone to data leakage issues plaguing other movie datasets (see Figure 2).

Aiming towards long-term story-level video understanding, we propose **SF20K-Test**, a manual Open-Ended Question Answering (OEQA) benchmark. Unlike restrictive multiple-choice formats, this task requires free-form answers, necessitating deeper narrative comprehension. All question-answer pairs in this benchmark are manually crafted and cover the setting, character dynamics, storylines, and themes of the films. To demonstrate the benefits of SF20K-Test, we conduct extensive experiments including a human study, baseline evaluation, and several ablation analysis.

Contributions are threefold: (1) We propose **SF20K**, the largest publicly available movie dataset, comprising 20,143 short films. We accompany this with **SF20K-Test**, a manually curated benchmark specifically designed to evaluate open-ended, story-level video understanding. (2) Through extensive experiments, we show: (i) Unlike other movie datasets, SF20K-Test exhibits limited data leakage to modern LLMs; (ii) State-of-the-art video models show significantly lower performance compared to humans; (iii) A long temporal window and a combination of vision and language are required to solve the QA tasks, highlighting the need for models capable of handling multimodal information at the movie level. (3) We demonstrate that synthetic instruction tuning on the large-scale SF20K dataset helps model's performance. We argue that SF20K provides a unique dataset for advancing long-term, story-level video understanding. The dataset, code, and models are publicly available at https://ridouaneg.github.io/sf20k.html.

## 2 Related Work

**Video QA benchmarks.** Questions in such benchmarks can be framed to assess the reasoning, memory, and comprehension skills of both humans and models. Hence, many datasets have been made available through the years, covering visual descriptions (Xu et al., 2016; Chen & Dolan, 2011), temporal action reasoning (Xiao et al., 2021), social intelligence (Zadeh et al., 2019), compositional reasoning (Grunde-Mclaughlin et al., 2021), instructional videos (Yang et al., 2021), movies (Tapaswi et al., 2016; Lei et al., 2019; Maharaj et al., 2017), egocentric videos (Mangalam et al., 2023; Jia et al., 2022; Lin et al., 2022), among others (Jang et al., 2017; Xu et al., 2017; Jang et al., 2017; Li et al., 2020; Yu et al., 2019; Wu et al., 2021). However, many benchmarks feature short videos and typically require reasoning on only a few frames to solve the task at hand, as shown by temporal certificates in Mangalam et al. (2023).

**Long-form video understanding.** As new long-context models emerge (Reid et al., 2024), we need more complex benchmarks. Long-form video understanding evaluates long-term reasoning capabilities of those models. Some datasets have been introduced in recent years, for general videos (Fu et al., 2024; Nagrani et al., 2024), egocentric videos (Mangalam et al., 2023), and movies (Wu & Krähenbühl, 2021; Soldan et al., 2022; Han et al., 2023; Tapaswi et al., 2016; Rawal et al., 2024b). EgoSchema (Mangalam et al., 2023) proposes multiple-choice questions on 3-minute videos that require on average 100 seconds of viewing time to be answered correctly. Video-MME (Fu et al., 2024) and Neptune (Nagrani et al., 2024) introduce longer videos that present significant challenges even for the most advanced vision-language models. On movies, LVU (Wu & Krähenbühl, 2021) focuses on clips, however, the subsequent tasks are relatively limited, focusing on less relevant aspects such as 'like' ratio and view count prediction. MAD/MAD-v2 (Soldan et al., 2022; Han et al., 2023) introduces an audio description task, which is currently tackled at the clip level but could be extended over the whole movie. MovieQA (Tapaswi et al., 2016) and MovieChat (Song et al., 2023) propose movie question-answering datasets focused on movie plots, settings, and characters, though access to the data is limited. Closer to our work, CinePile (Rawal et al., 2024b) offers a dataset of question-answer pairs derived from commercial movie clips, providing both training and test sets; however, similarly to other movie datasets, it suffers from data leakage issues.

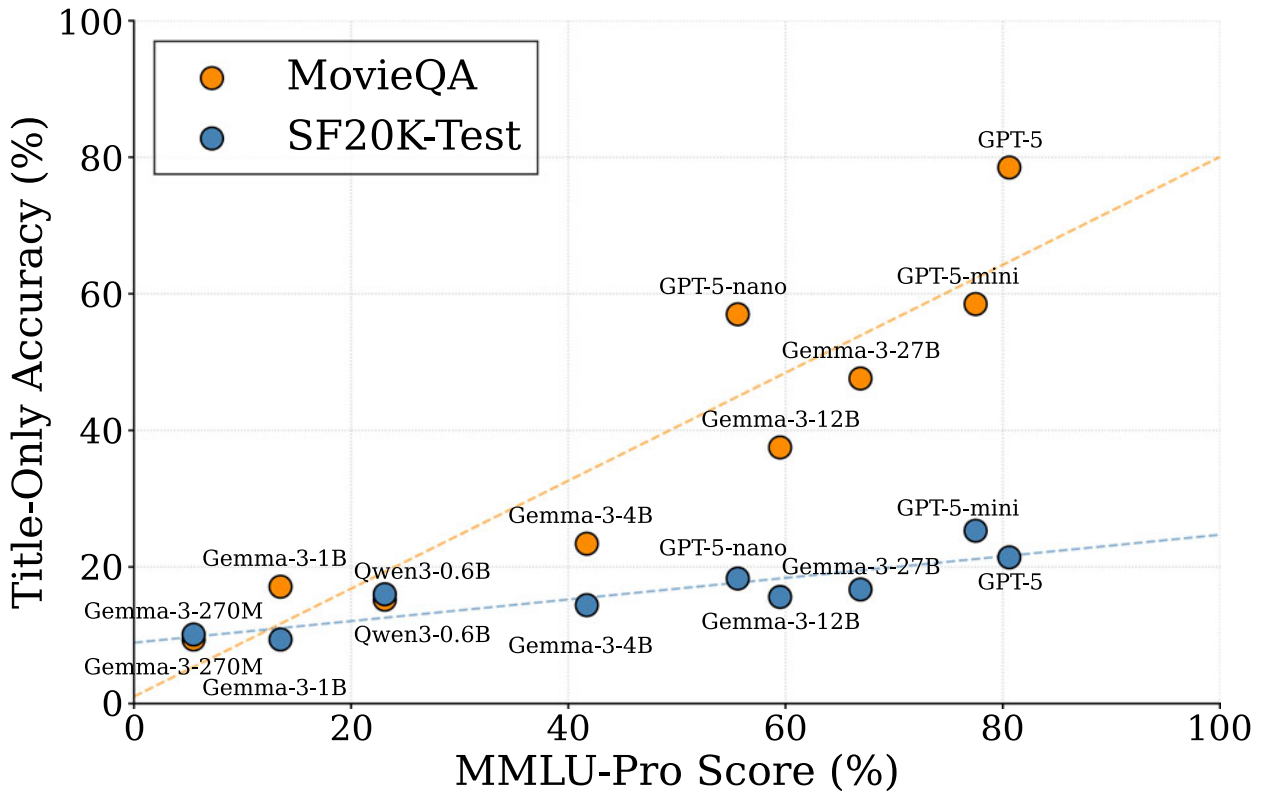

**Fig. 2** **Evaluating data leakage via title-only prompting.** LLMs show a sharp correlation between MMLU-Pro scores and the ability to "recall" commercial movie details without context (orange). In contrast, accuracy on unknown amateur films (blue) remains low, suggesting that high performance on popular datasets is driven largely by training data contamination rather than true understanding

## 3 The SF20K-Test Benchmark

*Short films* are motion pictures that typically last between 5 and 40 minutes. As illustrated in Figure 3, short films span a variety of styles (e.g., narrative fiction, documentary, animation) and genres (e.g., drama, comedy, horror). Filmmakers often use the short film format to explore new ideas, techniques, and storytelling styles. Short films, hence, have more of an experimental rather than a commercial value and are often made publicly available. The amount of publicly available short films has significantly increased over the last few years (see Figure 4b). We take advantage of this development and create SF20K, a dataset of short movies collected from online platforms. Among this dataset, we select high-quality movies to create SF20K-Test, a video question answering benchmark which includes manual, open-ended questions.

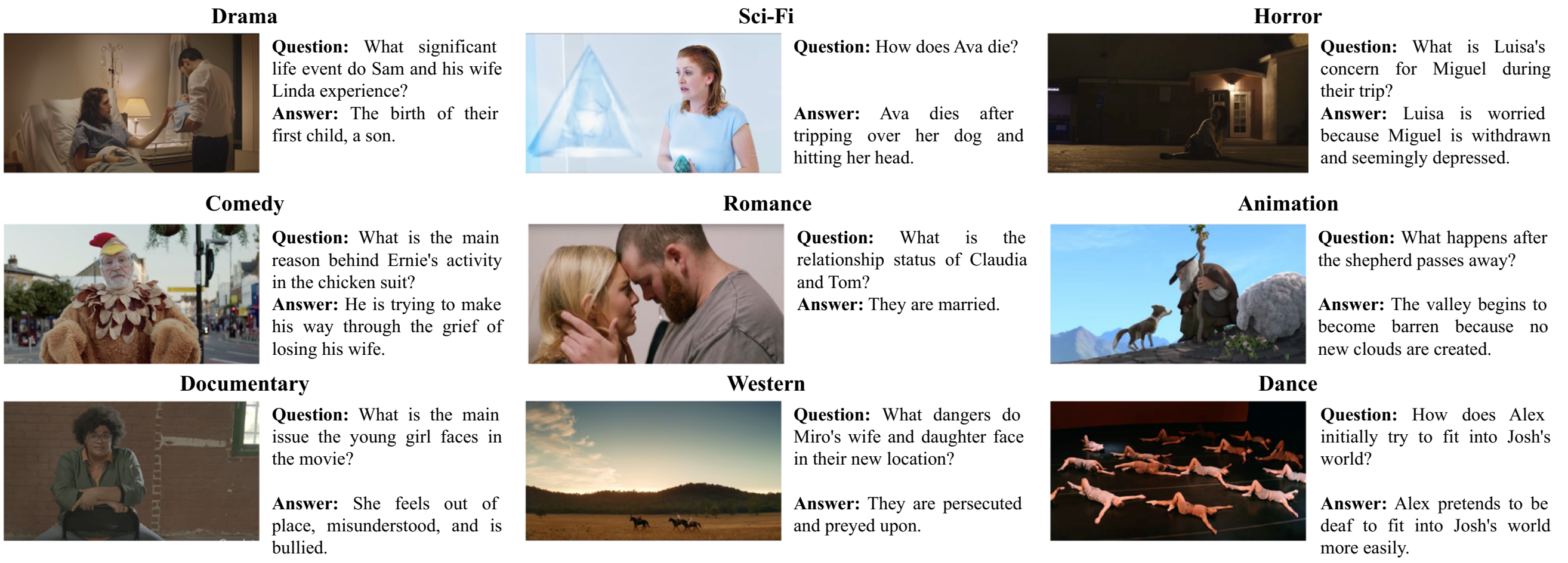

**Fig. 3 Examples from the SF20K dataset.** SF20K spans a wide range of genres with distinct visual styles and storytelling. For each genre, we show a representative frame together with a manually written question–answer pair from SF20K-Test, illustrating the story-level, character-driven nature of the benchmark

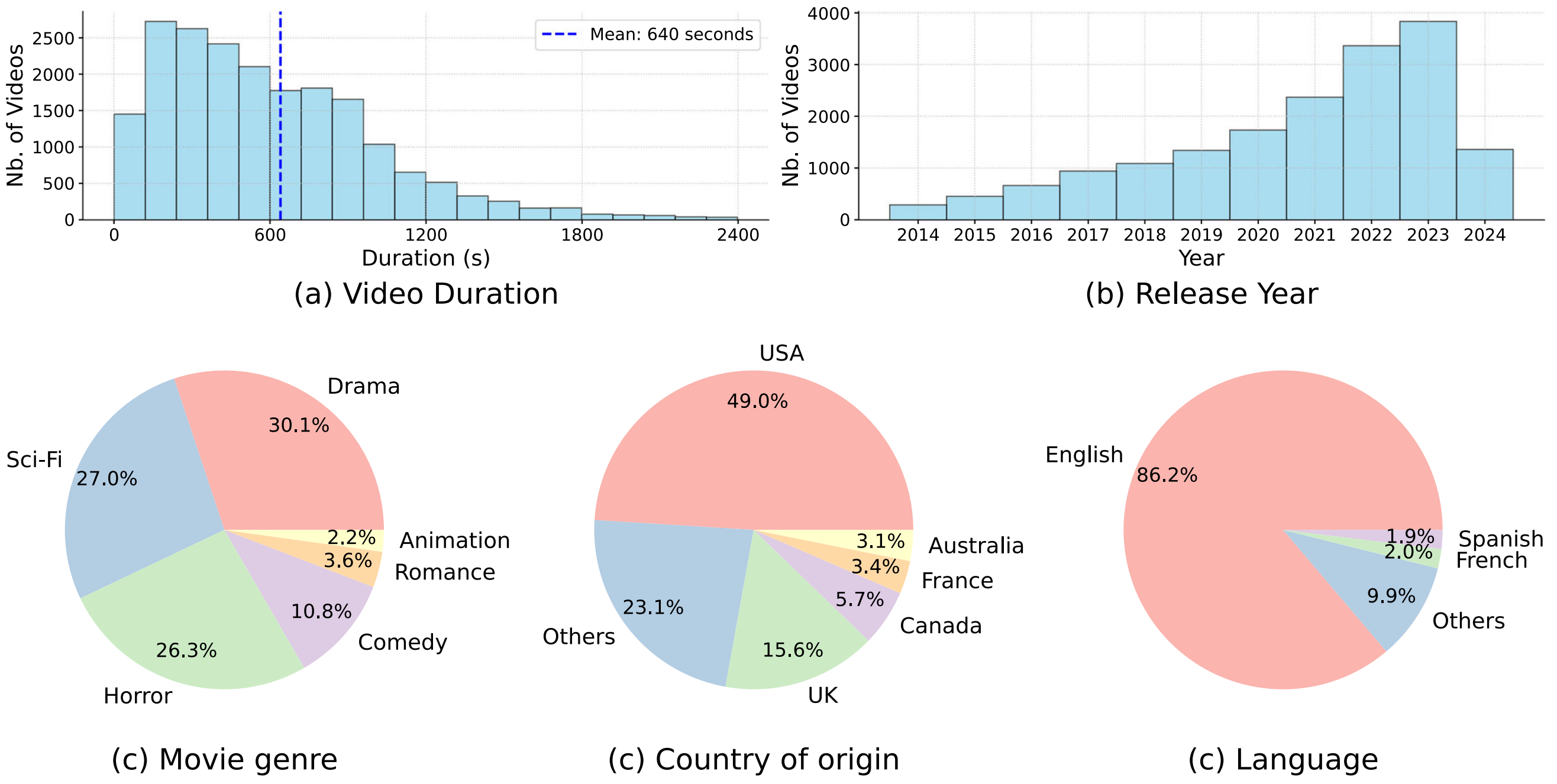

**Fig. 4 Statistics of SF20K**

## 3.1 Data Collection

To build SF20K, we capitalize on the abundance of short films on video-sharing platforms. We target YouTube and Vimeo channels hosting high-quality and award-winning short films, some of which have been recognized by Oscars and BAFTA awards [1]. Videos, subtitles, and metadata are downloaded using yt-dlp [2]. We filter out videos with fewer than 100 views or less than three months old (after December 2025), and rely on platform safety filters to exclude sensitive, age-restricted, and private content.

[1] Among others, we use the YouTube channels Omeleto https://www.youtube.com/@Omeleto, Film Shortage https://www.youtube.com/@FilmShortage, The Shorts Network https://www.youtube.com/@TheShortsNetwork, the Vimeo Staff Picks https://vimeo.com/channels/staffpicks, and the Shortverse library https://www.shortverse.com.

[2] https://github.com/yt-dlp/

### 3.2 Human Annotation

To construct a rigorous benchmark for story-level understanding, we recruited 10 human annotators to watch the full movies and generate 5-20 open-ended question-answer pairs per film. Annotators were instructed to focus on the narrative arc, character dynamics, settings, and key visual elements, ensuring that questions require reasoning rather than simple retrieval. To guarantee independence and clarity, guidelines specified that questions must explicitly refer to character names or specific plot moments and be answerable without external context. Both questions and answers were strictly curated to be concise, with answers limited to a single sentence. Typical questions were provided to the annotators. We provide several examples from the benchmark in Figure 3 and Figure S4.

**Human Performance.** To verify the quality of the human-created questions and establish a performance upper bound, we conducted a user study involving 10 other participants. Annotators were asked to watch the full movies and answer the question with a one-sentence, concise answer.

### 3.3 Benchmark Statistics

**Statistics.** SF20K-Test includes **95** movies, totaling **20.1** hours, with an average of **12.7** minutes per video (see Figure 5). This results in **979** question-answer pairs, averaging 10.3 questions per movie.

**Comparison to other benchmarks.** Table 2 compares SF20K-Test with existing video benchmarks. Notably, SF20K-Test exhibits a long average duration (762 seconds) among movie QA benchmarks. Compared to classic video QA benchmarks, SF20K-Test distinguishes itself by its multimodal and long-term focus. Compared to long-form video QA benchmarks, SF20K-Test focuses on complex narratives and character dynamics. Compared to movie QA benchmarks, SF20K-Test offers advantages in public availability of videos, limited data leakage (unlikely exposure to LLMs), and story completeness, as it uses full movies rather than trimmed clips, potentially impacting narrative coherence.

**Question Types.** To evaluate different dimensions of movie understanding, we categorize the questions into four types using an LLM (see Appendix G for the prompts used):

– *Setting-related questions:* Ask about the movie's time (historical period, time of day) and place (location, specific environments like home or office).
– *Character-related questions:* Focus on the main characters' personal details, traits, and their relationships.
– *Story-related questions:* Cover the key sequence of events and actions driving the main plot and narrative.
– *Theme-related questions:* Explore the central message, underlying idea, or abstract concept the movie conveys.

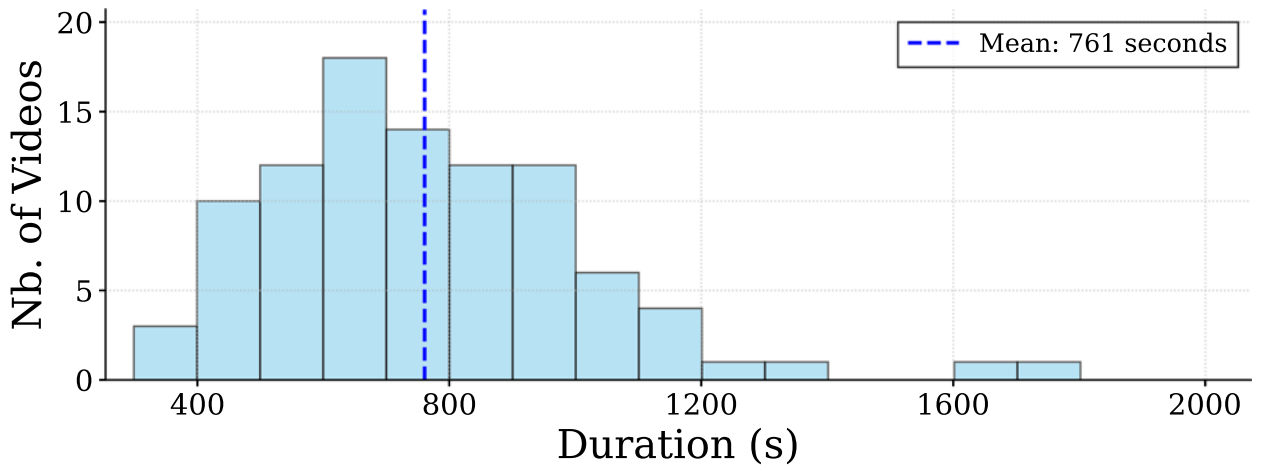

(a) Video Duration

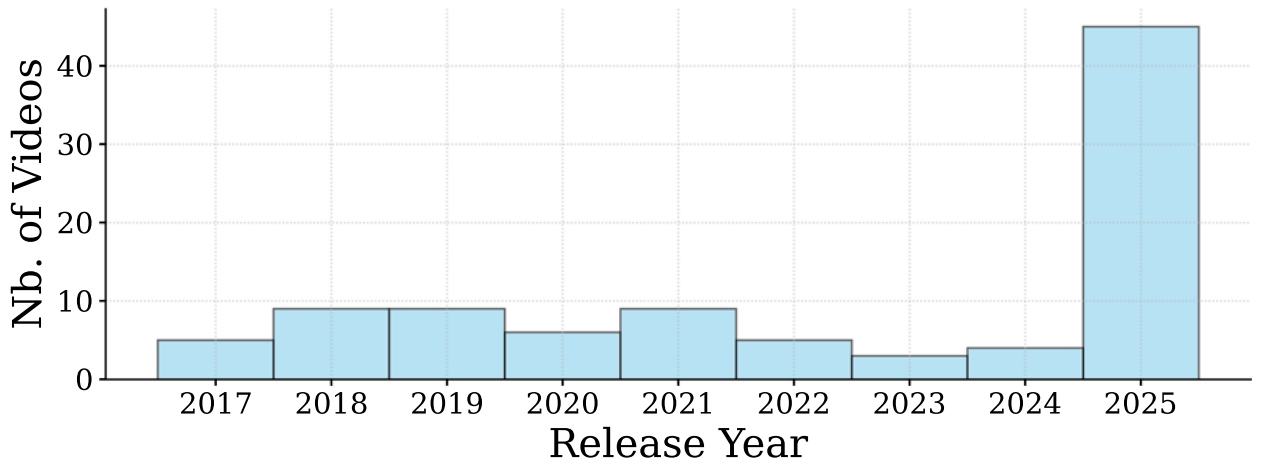

(b) Release Year

**Fig. 5** Statistics of SF20K-Test

### 3.4 Evaluation Metric

The open-ended nature of SF20K-Test makes traditional n-gram metrics (e.g., BLEU, ROUGE) unreliable, as they often penalize correct answers that use different phrasing. Instead, we utilize an LLM-as-a-judge approach (Zheng et al., 2023; Gu et al., 2025), which has demonstrated superior alignment with human judgment in complex reasoning tasks.

We define the **LLM-QA-Eval** metric: a judge model (GPT-4.1-nano) is provided with the original question, the human ground-truth answer, and the model's prediction. The judge is tasked with determining if the prediction is semantically consistent with the ground truth. We report the accuracy as our primary metric, defined as the percentage of "correct" labels assigned by the judge. The detailed evaluation prompt is provided in Appendix G and a human validation experiment in Section 5.2.

## 4 Large-scale QA generation

While the SF20K-Test benchmark serves as a high-quality, manual "gold standard" to evaluate story-level understanding, we extract annotations at scale on the full SF20K dataset to perform instruction tuning.

### 4.1 Automatic Generation

**Annotation Extraction** Since speech transcripts are available for only a portion of the dataset, we use Whisper Large-v3-Turbo (Radford et al., 2024) to generate miss-ing subtitles. The metadata accompanying each short film includes the movie title, storyline (a concise one-sentence plot summary), genre, release year, region of origin, and language. Additionally, we extract shot boundaries using PySceneDetect[3], face tracks using DeepFace[4], and frame captions using Llama-3.2-Vision-11B (Meta, 2022). More details can be found in Appendix B.

**Synopsis Generation.** We generate synthetic synopses by prompting GPT-4 (Openai, 2022) using the movie title, storyline, and extracted subtitles and captions. This process captures key story-level details, producing one synopsis per movie.

**QA pair generation.** From the synopses, we generate question-answer pairs using again GPT-4 (Openai, 2022). Prompts are carefully designed to obtain concise, specific, and unambiguous questions, with brief answers (see Appendix G).

### 4.2 Dataset Statistics

SF20K contains **20,143** short films ranging from 5 to 37 minutes, with an average length of **11** minutes per video (see Figure 4a), and totaling **3,582** hours. The dataset covers multiple genres, as illustrated in Figure 4. Examples of the dataset's diversity can be seen in Figure 3. Most of the films are in English and originate mainly from North America and Europe (see Figure 4(d,e)). More samples can be seen in Appendix S4.

**Annotations.** SF20K includes 2.4M shots, with an average shot length of 5.52 seconds. Each shot has a 24-word caption describing settings and characters. Additional annotations include 1.1M subtitles and 1.08M face tracks. Additionally, SF20K contains **191,007** question-answer pairs, averaging 9.21 questions per movie.

**Comparison to other datasets.** Table 1 compares SF20K to existing large-scale video datasets. General large-scale video datasets, such as WebVid-10M (Bain et al., 2022) and InternVid (Wang et al., 2024a), usually contain millions of short clips We propose a specialized dataset focused on movies. Among existing movie datasets, SF20K stands out as the largest film dataset ever released, consisting of 20,143 films for a total duration of 3,582 hours. This scale is comparable to large-scale specialized domain datasets like Ego4D (Grauman et al., 2022b) (3,670 hours). Furthermore, SF20K addresses key limitations of prior movie datasets: the videos are easily accessible online (unlike MovieNet (Huang et al., 2020b)), and the dataset contains full movies with complete stories (unlike Condensed Movies (Bain et al., 2020b) and TVQA (Lei et al., 2019)).

[3] https://github.com/Breakthrough/PySceneDetect

[4] https://github.com/serengil/deepface

## 5 Experiments

In this section, we introduce several experiments performed on the novel SF20K dataset. Section 5.1 examines data leakage in movie datasets. Section 5.3 benchmarks models on SF20K-Test. Section 5.4 analyzes model's performance across input modalities, number of frames, and temporal window. Section 5.5 analyzes the impact of instruction tuning on model's performance.

### 5.1 Data Leakage

Modern LLMs, pre-trained on vast internet data, risk being exposed during pretraining to information about movies, such as synopses, reviews, scripts, and subtitles. This exposure may result in models recalling answers directly without analyzing the video content, thereby inflating benchmark scores through a phenomenon known as *data leakage*. In this section, we quantitatively assess the extent of this bias by prompting LLMs to answer questions using *only* the movie title, evaluating the responses via our LLM-QA-Eval metric (see Section 3.4). We compare the results across two datasets: MovieQA (Tapaswi et al., 2016), which consists of commercial mainstream movies, and SF20K-Test, which focuses on amateur short films; and three models: Gemma-3 (Team et al., 2025), Qwen3 (Yang et al., 2025), and GPT-5 (Openai, 2025). See Appendix C for more details.

**Data leakage by LLM knowledge.** In Figure 2, we observe that the MovieQA benchmark suffers from data leakage, with LLMs reaching up to 79.2% accuracy using only the movie title as input. In contrast, thanks to the low presence of amateur films on the internet, SF20K-Test exhibits a maximum accuracy of 24.0%, indicating a low leakage issue. More importantly, the extent of data leakage is proportional to the knowledge level of the LLM, measured by MMLU-Pro score (Wang et al., 2024a), indicating that LLMs with more knowledge exhibit higher levels of memorization, leading to worse leakage issues. For instance, for MovieQA, the zero-shot accuracy increases from 10.2% to 79.2% as the LLM's knowledge increases. In contrast, SF20K-Test maintains stable and low accuracies ranging between 9.4% and 24.0%, regardless of LLM knowledge variations, further indicating limited knowledge of amateur movies.

To explicitly quantify data leakage, we propose a score defined as the sensitivity between an LLM's title-only accuracy (where it sees only the movie title and question) and its general knowledge (measured by MMLU-Pro). Figure 2 displays this score as the linear regression slope (computed across a wide range of LLMs) for MovieQA (orange) and SF20K-Test (blue). Figure S2 in Appendix provides results on more commercial TV serie/movie benchmarks (TVQA, CinePile, InfiniBench). As reported in Table 3, MovieQA's high score of 1.065 indicates that stronger LLMs can leverage parametric memory to ”know” the answers, whereas SF20K-Test results in a score of only 0.174, significantly lower than other movie benchmarks. This confirms that our benchmark, based on amateur films, presents limited data leakage compared to commercial movies.

**Table 1** Comparison of SF20K with large-scale video datasets

| Dataset | Venue | # Samples | Total Hours | Avg. Dur. (s) | Accessible Online | Full Story |
|---|---|---|---|---|---|---|
| *Web-Scale Clip Datasets* | | | | | | |
| WebVid-10M Bain et al. (2022) | ICCV 2021 | 10.7M | 42K | 14 | ✓ | - |
| HD-VILA-100M Xue et al. (2022) | CVPR 2022 | 103M | 760.3K | 11 | ✓ | - |
| InternVid-10M Wang et al. (2024a) | arXiv 2023 | 234M | 371.5K | 13 | ✓ | - |
| Panda-70M Chen et al. (2024) | CVPR 2024 | 70.8M | 166.8K | 8 | ✓ | - |
| MiraData Ju et al. (2024) | NeurIPS 2024 | 330K | 16K | 72 | ✓ | - |
| Koala-36M Wang et al. (2025a) | CVPR 2025 | 36M | 172K | 17 | ✓ | - |
| *Instructional Videos* | | | | | | |
| YouCook2 Zhou et al. (2017) | CVPR 2018 | 14,000 | 176 | 45 | ✓ | - |
| HowTo100M Miech et al. (2019b) | ICCV 2019 | 136M | 134,472 | 4 | ✓ | - |
| *Egocentric Videos* | | | | | | |
| Ego4D Grauman et al. (2022b) | CVPR 2022 | - | 3,670 | - | ✓ | - |
| EgoExo-4B Grauman et al. (2024) | CVPR 2024 | 5,035 | 1,286 | 919 | ✓ | - |
| *Instruction Tuning Datasets* | | | | | | |
| LLaVA-Hound Zhang et al. (2024d) | arXiv 2024 | 900,000 | 3,000 | 12 | ✓ | - |
| ShareGPT4Video Chen et al. (2024) | NeurIPS 2024 | 40,000 | 200 | 18 | ✓ | - |
| LLaVA-Video-178K Zhang et al. (2024e) | TMLR 2025 | 178,510 | 2,000 | 40 | ✓ | - |
| VideoMarathon Lin et al. (2025) | NeurIPS 2025 | 27,846 | 9,700 | 1,254 | ✓ | - |
| *Movies & TV Series* | | | | | | |
| TVQA Lei et al. (2019) | EMNLP 2018 | 21,793 | 460 | 76 | ✗ | ✗ |
| MovieNet Huang et al. (2020b) | ECCV 2020 | 1,100 | 3,000 | 9,818 | ✗ | ✓ |
| Condensed Movies Bain et al. (2020b) | ACCV 2020 | 33,976 | 1,270 | 134 | ✓ | ✗ |
| MovieChat-1K Song et al. (2023) | CVPR 2024 | 1,000 | 128 | 459 | ✓ | ✓ |
| CinePile Rawal et al. (2024a) | arXiv 2024 | 9,396 | 417 | 160 | ✓ | ✗ |
| InfiniBench Ataallah et al. (2025) | EMNLP 2025 | 1,217 | 1,066 | 3,155 | ✓ | ✓ |
| SeriesBench Zhang et al. (2025) | CVPR 2025 | 1,072 | 23 | 79 | ✗ | ✗ |
| **SF20K (Ours)** | - | 20,143 | 3,582 | 640 | ✓ | ✓ |

**Table 2** Comparison of SF20K-Test with Video QA benchmarks

| Dataset | Venue | Annotation | Avg. Dur. (s) | Total Hours | #Test Questions | Multimodal | Long Term | Accessible Online | Full Story | Question Type MCQ | Question Type Open |
|---|---|---|---|---|---|---|---|---|---|---|---|
| *General Videos* | | | | | | | | | | | |
| MSRVTT-QA Xu et al. (2017) | ACM 2017 | Auto | 15 | 41 | 72,820 | ✗ | ✗ | ✓ | - | ✓ | ✗ |
| MSVD-QA Chen and Dolan (2011) | ACM 2017 | Auto | 10 | 5 | 13,156 | ✗ | ✗ | ✓ | - | ✓ | ✗ |
| TGIF-QA Jang et al. (2017) | CVPR 2017 | Auto | 3 | 47 | 25,751 | ✗ | ✗ | ✓ | - | ✓ | ✗ |
| ActivityNet-QA Yu et al. (2019) | AAAI 2019 | Manual | 180 | 290 | 8,000 | ✗ | ✗ | ✓ | - | ✓ | ✗ |
| NeXT-QA Xiao et al. (2021) | CVPR 2021 | Manual | 44 | 66 | 17,742 | ✗ | ✗ | ✓ | - | ✓ | ✓ |
| *Instructional Videos* | | | | | | | | | | | |
| How2QA Sanabria et al. (2018) | EMNLP 2020 | Manual | 60 | 366 | 4,400 | ✗ | ✗ | ✓ | - | ✓ | ✗ |
| iVQA Liu et al. (2018) | ICCV 2021 | Manual | 18 | 50 | 10,000 | ✗ | ✗ | ✓ | - | ✓ | ✗ |
| *Egocentric Videos* | | | | | | | | | | | |
| EgoSchema Mangalam et al. (2023) | NeurIPS 2023 | Auto + Manual | 180 | 250 | 5,000 | ✗ | ✓ | ✓ | - | ✓ | ✗ |
| EgoTempo Plizzari et al. (2025) | CVPR 2025 | Auto + Manual | 45 | 6 | 500 | ✗ | ✗ | ✓ | - | ✗ | ✓ |
| *Long-Form Videos* | | | | | | | | | | | |
| Video-MME Fu et al. (2024) | arXiv 2024 | Manual | 1,017 | 254 | 2,700 | ✓ | ✓ | ✓ | - | ✓ | ✗ |
| Neptune Nagrani et al. (2024) | arXiv 2024 | Auto + Manual | 149 | 100 | 3,268 | ✓ | ✓ | ✓ | - | ✓ | ✗ |
| HourVideo Chandrasegaran et al. (2024) | NeurIPS 2024 | Auto + Manual | 2,742 | 381 | 12,976 | ✓ | ✓ | ✓ | - | ✓ | ✗ |
| LongVideoBench Wu et al. (2024) | NeurIPS 2024 | Manual | 473 | 494 | 6,678 | ✓ | ✓ | ✓ | - | ✓ | ✗ |
| LVBench Wang et al. (2024) | ICCV 2025 | Manual | 4,101 | 117 | 1,549 | ✓ | ✓ | ✓ | - | ✓ | ✗ |
| Minerva Nagrani et al. (2025) | ICCV 2025 | Manual | 720 | 44 | 1,515 | ✓ | ✓ | ✓ | - | ✓ | ✗ |
| MLVU Zhou et al. (2025) | CVPR 2025 | Auto + Manual | 1,730 | 446 | 3,102 | ✓ | ✓ | ✓ | - | ✓ | ✓ |
| VRBench Yu et al. (2025) | CVPR 2025 | Manual | 5,775 | 1,540 | 8,243 | ✓ | ✓ | ✓ | - | ✓ | ✗ |
| *Movies & TV Series* | | | | | | | | | | | |
| MovieQA (test) Tapaswi et al. (2016) | CVPR 2016 | Manual | 211 | 75 | 1,258 | ✓ | ✓ | ✗ | ✓ | ✓ | ✗ |
| TVQA (test) Lei et al. (2019) | EMNLP 2018 | Manual | 76 | 46 | 15,254 | ✓ | ✗ | ✗ | ✗ | ✓ | ✗ |
| CinePile (test) Rawal et al. (2024a) | arXiv 2024 | Auto + Manual | 160 | 7 | 4,941 | ✓ | ✓ | ✓ | ✗ | ✓ | ✗ |
| MovieChat-1K (test) Song et al. (2023) | CVPR 2024 | Auto + Manual | 459 | 22 | 14,000 | ✓ | ✓ | ✗ | ✓ | ✓ | ✗ |
| InfiniBench (test) Ataallah et al. (2025) | EMNLP 2025 | Auto + Manual | 3,155 | 106 | 8,770 | ✓ | ✓ | ✗ | ✓ | ✓ | ✓ |
| SeriesBench (test) Zhang et al. (2025) | CVPR 2025 | Auto + Manual | 79 | 2 | 2,919 | ✓ | ✗ | ✗ | ✗ | ✓ | ✗ |
| **SF20K-Test (Ours)** | - | Manual | 762 | 20 | 979 | ✓ | ✓ | ✓ | ✓ | ✗ | ✓ |

**Table 3** Data leakage score of different movie benchmarks

| Benchmark | Score (↓) |
|---|---|
| MovieQA Tapaswi et al. (2016) | 1.065 |
| TVQA Lei et al. (2019) | 0.550 |
| CinePile Rawal et al. (2024a) | 0.370 |
| InfiniBench Ataallah et al. (2025) | 0.423 |
| SF20K-Test | **0.174** |

**Data leakage by movie year.** To further verify data leakage, we hypothesize that the knowledge of LLMs about commercial movies depends on whether the movie release happened before or after the knowledge cutoff of the LLM (i.e., the date after which it no longer has access to new information from training data). To investigate this, we evaluate the GPT-4 and GPT-3.5 models on movies with different release dates. We first select the 50 most popular movies per year according to IMDb [5] and generate QAs from their IMDb synopses following the procedure in Section 4.1. We then evaluate the LLM's ability to answer questions given only movie titles. Figure 6 shows that GPT-4 and GPT-3.5 obtain over 80% accuracy for movies released before 2021; however, their performance drops notably for movies released after their respective knowledge cutoff [6]. On the contrary, both LLMs have low accuracy for SF20K-Test short films regardless of their release dates, further supporting our data leakage claims.

**Anonymization and shuffling.** To better understand the impact of movie titles on LLM performance, we perform two additional experiments.

**(i) Anonymization.** We examine the zero-shot performance of GPT-4 in three settings: (i) providing movie titles, (ii) removing movie titles, and (iii) removing movie titles with additional question anonymization. Question anonymization means replacing character names in the questions with generic placeholders, as this may help LLMs to recall specific movies (e.g., Harry Potter, James Bond). It enables us to measure relative accuracies, i.e. the performance improvement of an LLM if it can guess the movie. As shown in Table 4, removing movie titles causes a significant drop in performance for MovieQA (-17.4%), with further reduction after full anonymization (-39.9%). In sharp contrast, the performance on the SF20K-Test remains overall stable. This indicates that the ability to identify the movie is the main reason why LLMs perform so well in commercial movies in our data leakage experiment, while this does not apply to amateur films.

**(ii) Title Perturbation.** We examine data leakage in different scenario, where movie titles are replaced with random ones, either from commercial or amateur movies. As shown in Table 5, to give a commercial movie title as input causes a substantial drop in accuracy for both datasets, even falling below the “No Movie Title” baseline. We argue that commercial titles are known by the LLM and trigger the retrieval of memorized information. In contrast, providing amateur titles has a limited effect on performances, likely because the LLM does not recognize these titles.

These experiments show that (i) evaluating new methods on commercial movies can be misleading, and (ii) SF20K-Test provides a more reliable benchmark for story-level video understanding.

## 5.2 Metric validation

To validate the effectiveness of the LLM-QA-Eval metric, we conduct a pairwise human validation study. We present annotators with a question and two candidate responses and ask them to rank which one is more accurate relative to the ground truth. We then compute the Kendall $\tau$ correlation measure between the annotators' judgments and several metric' results. Annotators evaluate the same $N = 128$ set of response pairs. This multi-annotator approach allows us to measure inter-annotator agreement, which attains 0.880 (Krippendorff's $\alpha$).

The experimental results demonstrate that LLM-QA-Eval significantly outperforms traditional evaluation methods in mimicking human judgment. As shown in Table 6, standard metrics like CIDEr and BLEU-1 align with human rankings only approximately 50% of the time, a modest improvement over the 33% accuracy expected from random guessing. In contrast, our LLM-based metric achieves a much high agreement rate of 76.6%. Furthermore, with a Kendall's $\tau$ of 0.705, the data confirms that LLM-QA-Eval serves as a far more reliable proxy for human judgment for open-ended video question answering within the story-understanding domain.

## 5.3 Benchmarking SF20K-Test

In this section, we present a comprehensive evaluation of human and model performance on the SF20K-Test benchmark. Unlike previous movie benchmarks that rely on trimmed clips or synthetic questions, SF20K-Test requires

[5] https://www.imdb.com/search/title/?release_date=2024-01-01,2024-12-31

[6] Some recent commercial movies obtained high accuracy despite being released after LLMs. We found that most of them were either based on famous novels or older movies (‘Dune: Part Two’, ‘Wicked’); hence, they also suffer from data leakage.

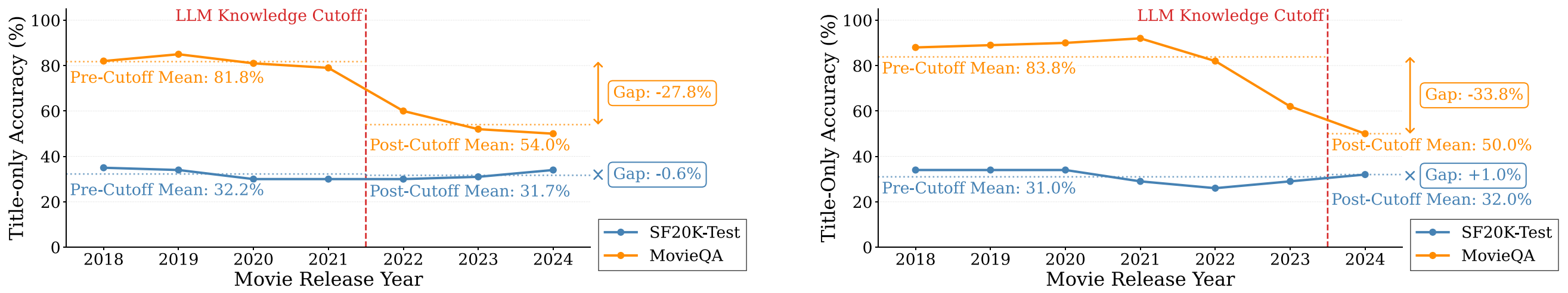

(a) GPT-3.5 (b) GPT-4

**Fig. 6** Data leakage by year. Title-only accuracy of GPT-3.5 and GPT-4 on MovieQA (orange) and SF20K-Test (blue). The vertical lines mark the LLMs' knowledge cutoffs. The gap between pre- and post-cutoff performance on MovieQA indicates heavy data leakage. In contrast, SF20K (amateur) accuracy remains consistently low and stable across all years, demonstrating a lack of LLM pre-training exposure

**Table 4** Impact of anonymization on data leakage. For MovieQA, the ability to identify the movie is crucial for the LLM to get a high accuracy

| Anonymization | Description | MovieQA | $+\Delta$ | SF20K-test | $+\Delta$ |
|---|---|---|---|---|---|
| No | - | 71.3 | | 18.3 | |
| Partial | Title Removed | 53.9 | **-17.4** | 17.1 | **-1.2** |
| Full | Title Removed & Names Anonymized | 31.4 | **-39.9** | 15.9 | **-2.4** |

**Table 5** **Impact of movie title perturbations on data leakage.** Providing known commercial movie titles misleads the LLM, resulting in a significant accuracy drop on both datasets. In contrast, random amateur titles have minimal impact.

| Input Scenario | Commercial (MovieQA) | Amateur (SF20K-Test) |
|---|---|---|
| Correct Title | 57.0 | 18.3 |
| Random Commercial Title | 20.3 | 8.5 |
| Random Amateur Title | 32.0 | 16.7 |
| No Movie Title | 36.3 | 17.1 |

**Table 6** Correlation of metrics with human judgments ($N = 128$). All Kendall $\tau$ correlations are significant at $p < 0.05$

| Metric | % Agree [0, 100] | Kendall $\tau$ [−1, 1] |
|---|---|---|
| Random | 33.3 | 0.000 |
| BLEU | 45.3 | 0.313 |
| CIDEr | 50.0 | 0.353 |
| ROUGE | 51.6 | 0.372 |
| BERTScore | 51.6 | 0.445 |
| LLM-QA-Eval | **76.6** | **0.705** |

reasoning over full-length narratives (avg. 12.7 min) and complex character arcs.

**Human performance.** Section 3.2 details the human study. As reported in Table 7, human annotators achieve an accuracy of 91.7%, which demonstrates that questions are well-formed and answerable, and sets an upper bound for models to reach. A score below 100% can be attributed to three main sources of error: (1) human mistakes, (2) subjective answer, and (3) inaccuracies in the metric computation.

**Baseline performance.** We evaluate recent open-source and commercial vision-language models on SF20K-Test. In particular, we benchmark 13 open-source models: Video-LLaVA (Lin et al., 2024a), Llava-OneVision (Li et al., 2024b), Llama-3.2-Vision (Meta, 2022), Pixtral (AI, 2024a), Long-LLaVA (Wang et al., 2025), LongVA (Zhang et al., 2024b), LongVU (Shen et al., 2025), LLoVi (Zhang et al., 2024a), InternVL3.5 (Wang et al., 2025b), LLaVA-Video (Zhang et al., 2025), LVAgent (Chen et al., 2025), Qwen2.5-VL (Bai et al., 2025b), and Qwen3-VL (Bai et al., 2025a); and 2 commercial models: GPT-5 (Openai, 2025), and Gemini-2.5 (Google, 2025). See Appendix E for more details on prompts and metrics. Models are tested in a zero-shot video question-answering setting, adapted for long-form video understanding by incorporating subtitles and sampling the maximum number of frames depending on the model's capacity. For open-source models, we sample between 8 (for older models) and 256 frames (for most recent ones). For proprietary models, we maximize the number of frames within API limits (i.e., 1 FPS for Gemini-2.5-Pro and 500 frames for GPT-5). We evaluate performance using the LLM-QA-Eval metric (see Section 3.4), which uses an LLM-as-a-judge to determine the semantic similarity between model predictions and human ground-truths. The final score is the accuracy (i.e., percentage of correct answers).

As shown in Table 7, the SF20K-Test benchmark proves highly challenging for current models. Gemini-2.5-Pro leads the benchmark with an accuracy of 59.3%, while the best-

**Table 7** Zero-shot performance on SF20K-Test

| Model | #Frames | Accuracy |
|---|---|---|
| Random | - | 0.4 |
| Question-only | - | 14.7 |
| *Open-source models* | | |
| Video-LLaVA-7B Lin et al. (2024b) | 8 | 3.5 |
| Llava-OneVision-0.5B Li et al. (2024a) | 8 | 19.5 |
| Llava-OneVision-7B Li et al. (2024a) | 8 | 29.0 |
| LLoVi Zhang et al. (2024a) | 256 | 26.1 |
| Long-Llava-7B Wang et al. (2025) | 8 | 29.5 |
| Llama-3.2-Vision-11B AI (2024b) | 8 | 31.0 |
| InternVL3.5-8B Wang et al. (2025b) | 32 | 31.0 |
| InternVL3.5-14B Wang et al. (2025b) | 32 | 35.2 |
| LongVA-7B Zhang et al. (2024c) | 256 | 31.1 |
| Qwen2.5-VL-3B Bai et al. (2025b) | 32 | 31.7 |
| Pixtral-12B AI (2024a) | 8 | 33.0 |
| LLaVA-Video-7B Zhang et al. (2025) | 32 | 37.0 |
| LVAgent Chen et al. (2025) | 32 | 37.9 |
| LongVU-7B Shen et al. (2025) | 256 | 39.9 |
| Qwen3-VL-8B Bai et al. (2025a) | 256 | 50.1 |
| Qwen3-VL-32B Bai et al. (2025a) | 256 | 53.9 |
| *Proprietary models* | | |
| GPT-5-nano Openai (2025) | 500 | 41.0 |
| GPT-5-mini Openai (2025) | 500 | 52.2 |
| GPT-5 Openai (2025) | 500 | 56.2 |
| Gemini-2.5-Pro Google (2025) | 1 FPS | 59.3 |
| Human | - | 91.7* |

* human performance is assessed on 47.3% of the test set

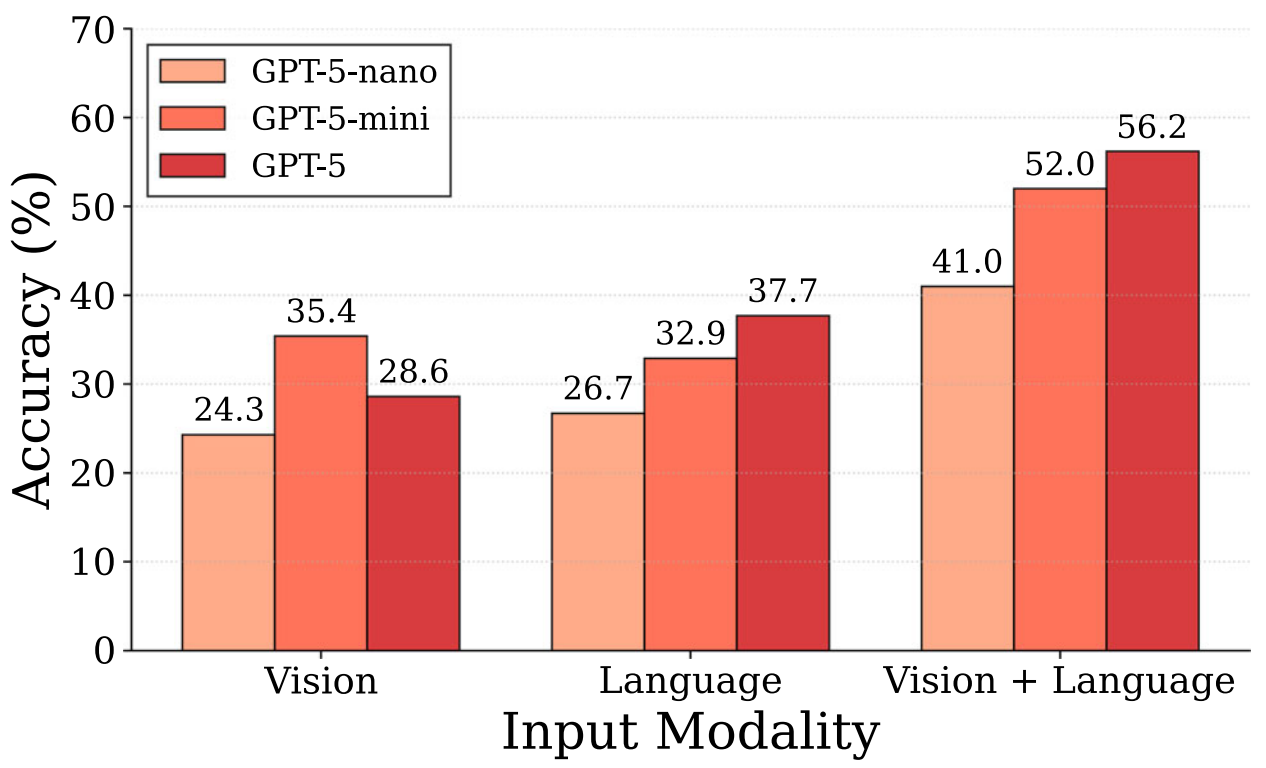

**Fig. 7** **Accuracy vs. Input Modality** of GPT-5 variants on SF20K-Test

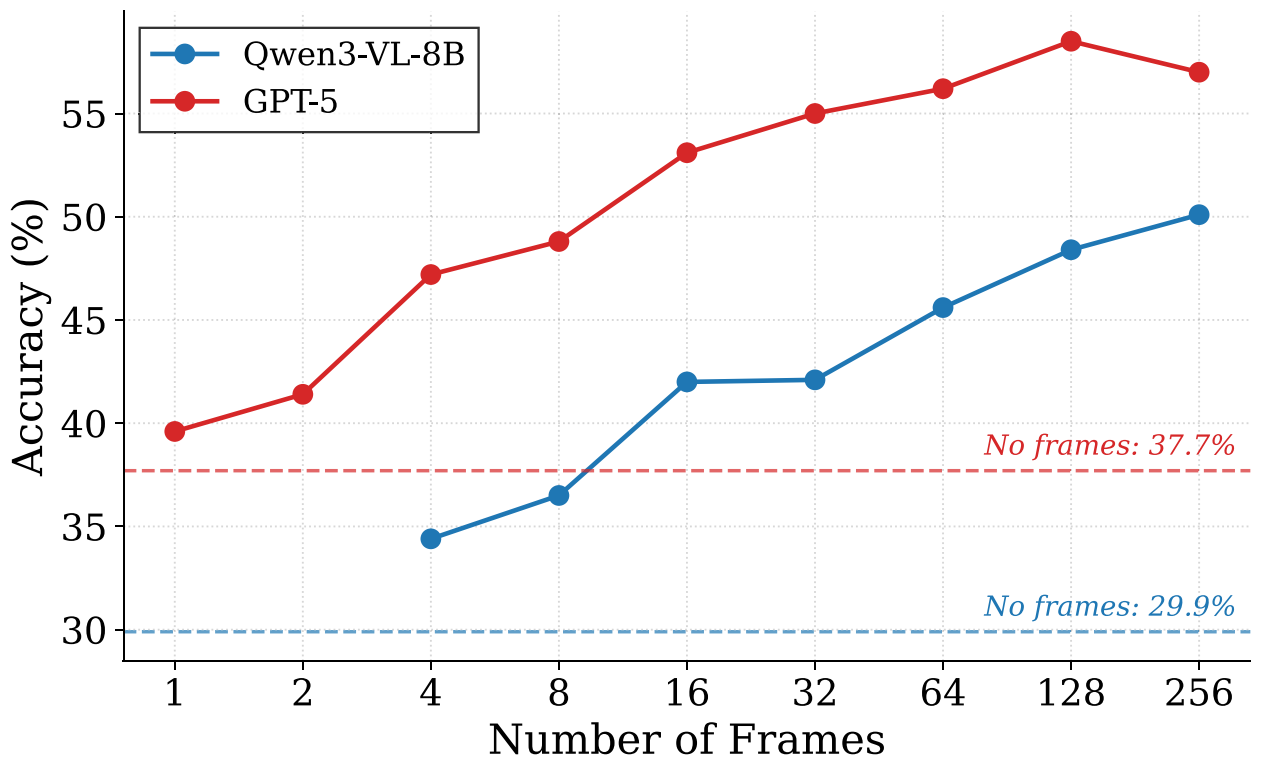

**Fig. 8** **Accuracy vs. Number of Frames** of GPT-5 and Qwen3-VL-8B on SF20K-Test

performing open-source model, Qwen3-VL-32B, reaches 53.9%. These results highlight a significant gap (over 32 points) compared to human performance, suggesting that even state-of-the-art models struggle with long-form story understanding.

**Failure case analysis.** We provide an analysis of failure cases to better understand the limitations of recent models in story-level video question answering. We identify three core challenges: (i) Character identification: The ability to identify the main characters, assign names, and recognize them in subsequent scenes; (ii) Long-range causal reasoning: The ability to connect distant scenes to keep track of the overall story; (iii) Visual story understanding: The ability to perceive key visual elements that are required to understand the story. We illustrate each one of these failure cases with qualitative examples in Figure S4.

### 5.4 Ablations

**Input modality.** In Figure 7, we evaluate model performance with different input modalities: *Vision* using video frames, *Language* with text subtitles, and *Vision + Language* combining both modalities. Multimodal performance consistently surpasses unimodal performances, indicating that recent models successfully integrate diverse signals to improve narrative understanding. Specifically, for GPT-5, accuracy increases from 28.6% in the Vision setting and 37.7% in the Language setting to 56.2% when both modalities are provided. While the multimodal setting yields the highest results, Language remains the dominant modality, as evidenced by higher accuracy in Language settings compared to Vision settings for most variants. However, while subtitles provide a strong signal, visual context is essential and complementary for resolving the complex story-level dynamics inherent in SF20K-Test.

**Number of frames.** In Figure 8, we explore model performance with respect to the number of input frames using GPT-5 and Qwen3-VL-8B. Accuracy consistently improves for both models as the number of frames increases, highlighting the importance of temporal resolution for story-level understanding. Notably, while many existing video benchmarks exhibit a "single-frame bias", where a single image provides enough context to answer the majority of questions, SF20K-Test requires significantly more temporal density to resolve complex narrative arcs. For instance, GPT-5 improves

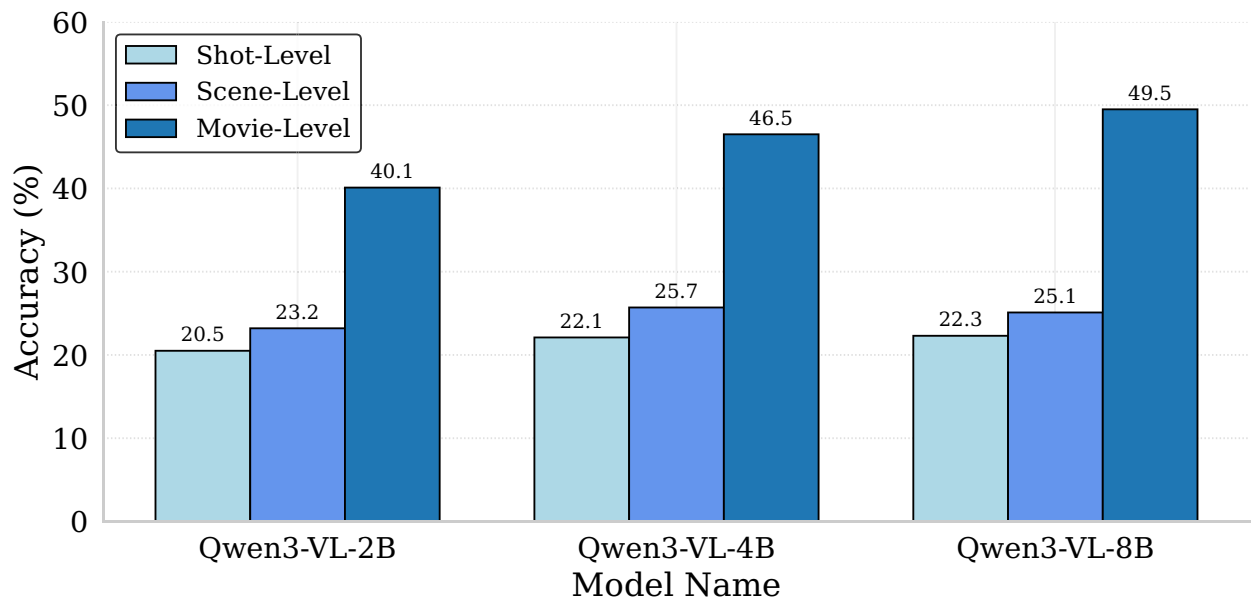

**Fig. 9** **Temporal window study.** We report the accuracy of Qwen3-VL across different temporal windows (shot, scene, movie) and model sizes (2B, 4B, 8B) on SF20K-Test

from 39.6% with a single frame to a peak of 58.6% at 128 frames, significantly outperforming its "No frames" (Language-only) baseline of 37.7%.

**Temporal window study.** To confirm that our tasks require long-form video understanding, we conduct a temporal window study by performing inference at three levels: shot-, scene-, and movie-level. Shots, defined as continuous video clips, last 5.52 seconds on average. At the *Shot-Level*, the model uses data from one shot, including partial subtitles and video frames. Predictions are aggregated by using an LLM picker, which selects the most probable answer out of the local predictions given the question as context. *Scene-Level* inference aggregates data from 10 shots, in a similar manner. *Movie-Level* inference uses the entire film's data. This study is conducted on SF20K-Test with Qwen3-VL using three model sizes: 2B, 4B, and 8B. More details in Appendix F.

Our results, depicted in Figure 9, reveal that all model sizes show substantial improvements with larger temporal windows. For instance, Qwen3-VL-8B's accuracy increases from 22.3% at the shot-level to 25.1% at the scene-level (+2.8%), to 49.5% at the movie-level (+27.2%). In conclusion, larger temporal windows substantially enhance the performance on SF20K-Test. This underscores the long-term nature of our proposed story-level video understanding task.

## 5.5 Instruction tuning

In this experiment, we fine-tune Qwen2.5-VL-3B (Bai et al., 2025b) on SF20K (Section 4.1). The inputs include the question, the video frames, and the subtitles extracted from the video. We use next-token prediction to generate responses. Fine-tuning is performed over $10^5$ samples from 19,676 movies for 1 epoch, with a learning rate of 1.0e-5, 375 warmup steps, a batch size 8, and gradient normalization of 1.0. We only fine-tune the LLM using LoRA (1.28% of the parameters), while keeping the vision encoder and the projection layer frozen. Samples are randomly presented as open-ended questions.

**Table 8** Performance improvement after fine-tuning Qwen2.5-VL-3B on SF20K. Performance is evaluated on SF20K-Test

| Model | #Frames | Fine-tuned | Accuracy | $+\Delta$ |
|---|---|---|---|---|
| Qwen2.5-VL-3B | 32 | ✗ | 31.7 | |
| Qwen2.5-VL-3B | 32 | ✓ | 35.6 | *+3.9* |

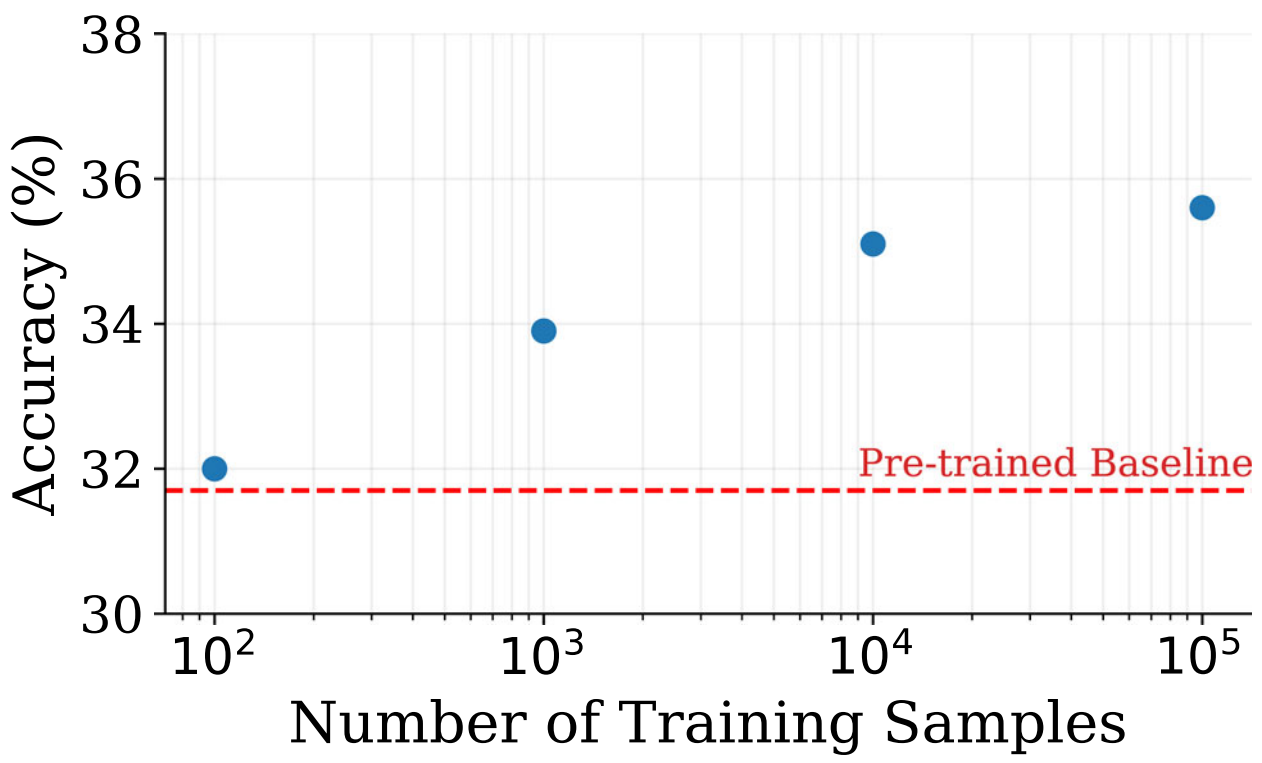

**Fig. 10** **Effect of the training set size** on the fine-tuning of Qwen2.5-VL-3B on SF20K

**Instruction tuning performance.** Table 8 reports the results, showing that instruction tuning enhances performance on SF20K-Test compared to the pretrained model. With 32 frames as input, Qwen2.5-VL-3B increases from 31.7% to 35.6% accuracy (+3.9%).

**Effect of the training size.** Figure 10 shows the performance of Qwen2.5-VL-3B (Bai et al., 2025b) when instruction-tuned for open-ended question answering on SF20K, with varying numbers of training samples. We subsample the full dataset at incremental values. Results indicate a logarithmic improvement in accuracy as the training set size increases. Results are displayed in logarithmic scale. These findings highlight the scalability benefits of instruction tuning on SF20K.

# 6 Discussion

We introduced Short-Films 20K (SF20K), a long-form video understanding dataset featuring 20,143 amateur movies. It includes open-ended questions, automatically generated by LLMs and, in the test set, manually created by human annotators. Compared to other video datasets, SF20K stands out by offering richer, story-oriented content with tasks specifically tailored for long-term reasoning, as validated by our experiments. Unlike other movie datasets that use copyrighted commercial movies prone to data leakage with LLMs, SF20K relies on amateur films, which are publicly accessible and have limited online presence. Our analysis indicates that state-of-the-art multimodal methods fall behind human accuracy, implying room for improvement and the need for further advancement of vision-language models. Finally, we show that instruction-tuning VLMs on SF20K-Train improves performance. In conclusion, we believe that SF20K paves the way for accessible and comprehensive movie understanding that is not known a priori by LLMs, thus helping the community to develop robust methods.

**Supplementary Information** The online version contains supplementary material available at https://doi.org/10.1007/s11263-026-02940-x.

**Acknowledgements** This work was supported by ANR-22-CE23-0007, a Hi!Paris grant and fellowship, and the ANR/France 2030 program (ANR-23-IACL-0005). Additional support was provided by the Institute of Information & Communications Technology Planning & Evaluation (IITP), funded by the Korean Government (MSIT) under Grant No. RS-2024-00457882 (National AI Research Lab Project). This work was granted access to the High-Performance Computing (HPC) resources of IDRIS under the allocations 2023-AD011014489 made by GENCI. We would like to thank Nicolas Dufour, Robin Courant, and Pierre Vassal for their insightful comments and suggestions. We also express our gratitude to the annotators of the human study for their help.

**Data Availability** The SF20K dataset is publicly accessible on the HuggingFace hosting platform. (https://huggingface.co/datasets/rghermi/sf20k)

**Publisher's Note** Springer Nature remains neutral with regard to jurisdictional claims in published maps and institutional affiliations.

## Appendix

In the following, we present the items:

A. Ethical considerations
B. Data pre-processing and statistics
C. Technical details on the data leakage experiment
D. Human study
E. Technical details on baselines
F. Technical details on the temporal window analysis
G. Prompts

## A Ethical considerations

1. Copyright and License:
   (a) We do not distribute raw video content; instead, we provide only URLs redirecting to YouTube and Vimeo, where the full copyright and licensing rights of creators are acknowledged. Additionally, by sharing the URLs, YouTube's and Vimeo's Terms of Service regarding exclusive rights are respected.[8].
   (b) The dataset is intended exclusively for academic research purposes. To ensure the protection of the dataset and all its associated elements, we use the Creative Commons Attribution-NonCommercial-ShareAlike 4.0 International (CC BY-NC-SA 4.0) license.
2. Potential negative social impact:
   (a) **Bias:** Collected movies are primarily from North America and Europe, mainly English-speaking, the under-representation of the content towards non-English-speaking population and culture, potentially leading to unbalanced, biased video understanding model training and testing.
   (b) **Impact on creative process:** Progress on movie understanding may encourage filmmakers to tailor their work to optimize for what is favoured by algorithms, potentially limiting creative freedom and depriving innovation.

## B Data pre-processing and statistics

SF20K provides not only videos but also a comprehensive set of annotations to facilitate deeper analysis and understanding. These annotations include subtitles, shot boundaries, keyframe captions, and face tracks. Together with tools for downloading, pre-processing, and annotating the data, these resources are released as part of the dataset. Below, we detail the processes involved in creating these annotations.

**Shot boundaries.** In the context of movies, a shot is a continuous sequence of frames captured without interruption by a single camera, often representing a single perspective or action within the scene. This corresponds to a clip in other video datasets. We use PySceneDetect[9] with a threshold equal to 3.0 to segment the 20,143 movies in SF20K into 2,396,583 shots, with each movie containing an average of 119 shots.

**Keyframe captions.** For each shot, a detailed caption is generated by extracting the middle frame of the shot and processing it with Llama-3.2-Vision [58]. The model is specifically prompted to identify and describe relevant details, including the time, location, character appearances, mood, and overall atmosphere. We end up with one caption per shot, i.e. 2,396,583 captions, for a total of 59,760,112 words. These textual annotations provide a good basis for further work on video generation models.

[8] `https://www.youtube.com/static?template=terms` (Section: License to Other Users), `https://vimeo.com/terms` (Section: Licenses Granted by You)

[9] https://www.scenedetect.com/

| | #Movies | #Hours | #QA Pairs | Source | Annotation |
|---|---|---|---|---|---|
| SF20K | 20,143 | 3,584 | 191,007 | Subtitles | Auto |
| SF20K-Test | 95 | 20 | 979 | Movie | Manual |

Table S1: **Detailed statistics of SF20K train and test sets.**

| | SF20K |
|---|---|
| #Movies | 20,143 |
| #Shots | 2,396,583 |
| #Shots / Movie | 119 |
| Avg. Shot Len. (s) | 5.52 |
| #Shot Captions | 2,396,583 |
| #Subtitles | 1,088,140 |
| #Face Tracks | 1,083,850 |

Table S2: **Additional statistics.** Statistics for the shots and face tracks in all movies of SF20K.

**Subtitles.** We extract missing subtitles from the audio with Whisper Large-v3 Turbo [66], using the translation mode. We end up with 1,088,140 dialog sentences, for a total of 12,038,891 words.

**Face tracks.** We use a facial recognition pipeline [10] to identify over 1,083,850 face tracks. We first detect faces with RetinaFace [21], and then compute face tracks by matching face detections among consecutive frames of the same shot with IoU matching. These annotations enrich the dataset, providing deeper insights for tasks related to character analysis and interaction within films.

**Word cloud analysis.** In Figure S1, we perform a word cloud analysis for three video domains commonly used in the literature: instructional videos [101], egocentric videos [56], and movies (ours). We notice that instructional and egocentric videos are focused on actions (verbs like make, go, cut, use, pick, put) and objects (e.g. table, sugar, oil, thing, stuff, table, paper), with an emphasis on body parts for egocentric videos (right hand, left hand). For movies, the vocabulary is more focused on characters (e.g. man, woman, boy, girl, young, old), relationships (e.g. mother, father, family, couple), time (e.g., day, night, year), and locations (e.g., home). These differences highlight the story-centric nature of the SF20K dataset.

**Subset statistics.** In Table S1, we provide the key statistics of SF20K train and test sets. One key element to highlight is the annotation process. SF20K comes with almost 200K question-answer pairs, used for instruction tuning. Those have been generated automatically from subtitles. On the contrary, questions from SF20K-Test is a fully manual test set, where questions are designed by annotators who watched the movies.

**Question types.** We use an LLM to automatically classify the question into 4 categories: setting-, character-, story-, and theme-related questions. In Table S3, we provide a few examples of questions for each question type from SF20K-Test.

## C Technical details on the data leakage experiment

In this section, we include additional information on the data leakage study mentioned in Section 5.1. The objective of our data leakage study is to demonstrate that current large language models (LLMs) possess prior knowledge about commercial movies. This is because LLMs have been exposed to vast amounts of movie-related textual information during pretraining, including synopses, blog posts, news articles, and more. The experiment is set up as an *open-ended question-answering*

[10] https://github.com/serengil/deepface

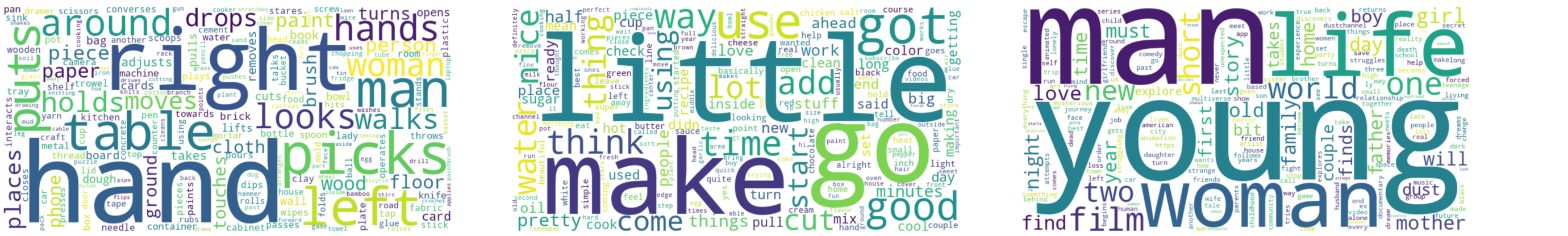

(a) Instructional videos (iVQA) (b) Egocentric videos (EgoSchema) (c) Movies (SF20K-Test)

Fig. S1: **Comparative word cloud analysis** across video domains: instructional videos (iVQA), egocentric videos (EgoSchema), and movies (SF20K-Test).

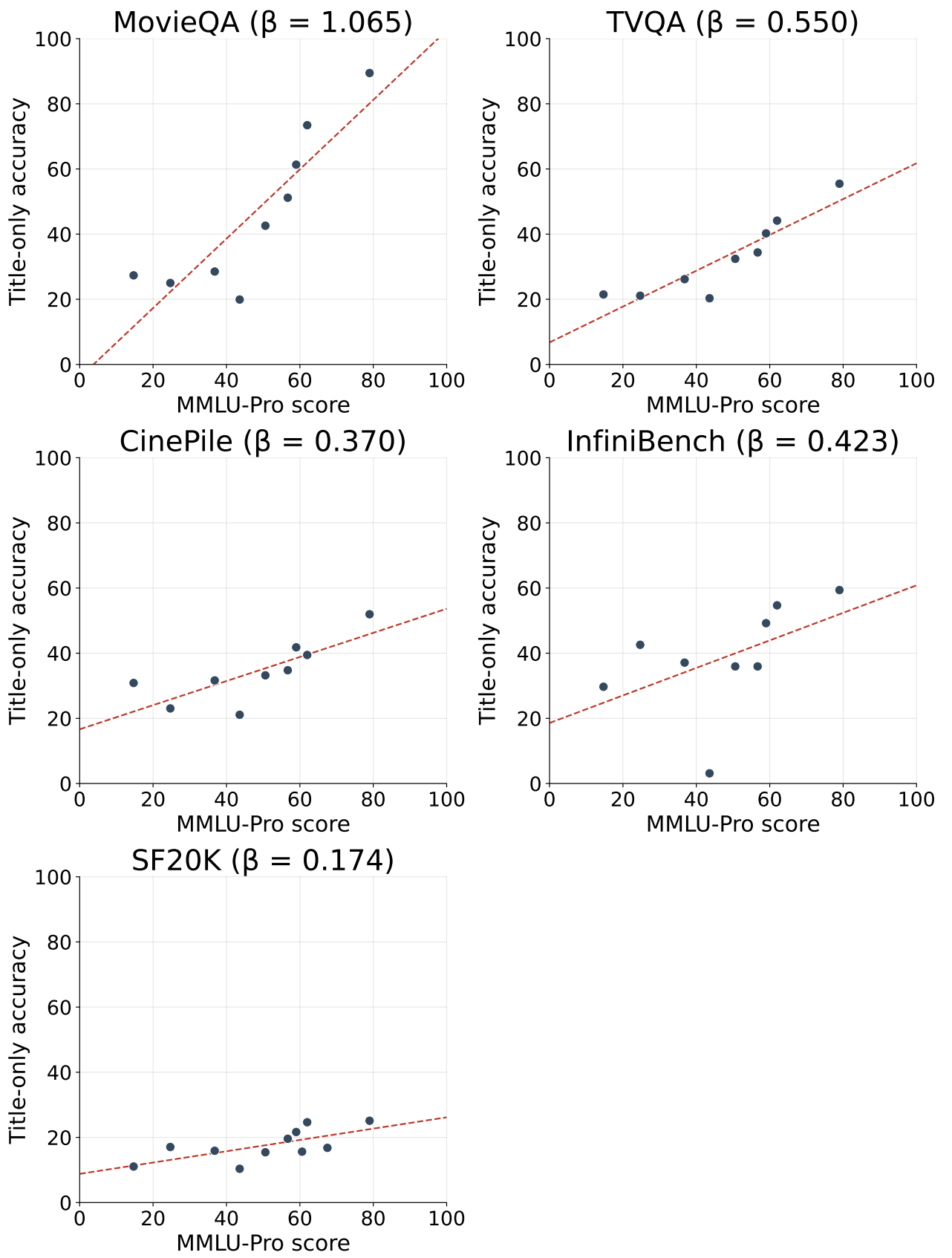

Fig. S2: **Data leakage scores** correspond to the sensitivity of the title-only accuracy to the LLM's general knowledge (measured by the MMLU-Pro score).

task where the input consists of the movie title and corresponding question (see Table S4 for a few examples), and the output is a free-form text generated by the LLM. We evaluate the answer using an LLM to compute the similarity between the predicted and the correct answer, following [54].

We choose recent LLMs: Qwen3, Gemma-3, and GPT-5. The open-source models are run using the Hugging Face transformers library [11]. The commercial models are run using their official API [12].

[11] https://github.com/huggingface/transformers

[12] https://openai.com/api

We choose two movie datasets: MovieQA [80], and SF20K-Test (Ours). For MovieQA and SF20K-Test, the question-answer pairs are directly available.

## D Human Study

In this section, we present additional details for the conducted user study (see Section 5.3). We conducted a comprehensive user study to evaluate the upper-bound performance on our benchmark. Participants were asked to watch a movie and subsequently answer multiple-choice questions based on the content. To analyze the impact of different modalities, we implemented two ablation conditions: a muted version, where participants had access to the vision-only information without audio or subtitles, and a blind version, where participants were provided with the audio-only information, in the form of textual subtitles, without any video.

To avoid bias, we ensured that different sets of participants were assigned to different settings (full, vision-only, and audio-only), preventing any overlap and potential influence from experiencing multiple conditions. Movies were randomly assigned to users to ensure a diverse range of data. The entire study was facilitated through an integrated interface, which restricted access to external information such as movie loglines and synopses, with only the movie title being displayed (see Figure S5).

## E Technical details on baselines

In this section, we present additional details for the baselines discussed in Section 5.3, where we report the benchmark performance of the state-of-the-art video understanding methods on SF20K-Test. We detail the models evaluated, the prompting strategy, and the metrics used.

**Models.** We evaluate nine open-source models: Video-LLaVA [49], Llava-OneVision [45], Llama-3.2-Vision [58], Pixtral [2], Long-LLaVA [87], LongVA [108], LongVU [75], Qwen2.5-VL [6], and Qwen3-VL [5]; and two commercial models: GPT-5 [64], and Gemini-2.5 [27].Most open-source models are available on Hugging Face's transformers library [13], except for LongVA [14], and LongVU [15]. Commercial models are available through their official API. Note that commercial models are not free; therefore, we are limited in their usage.

**Prompting.** We prompt all the models with the same queries, inspired by Gemini [70]. For open-ended question answering, we use the following prompt:

*'You will be given a question about a movie. Try to answer it based on the subtitles and the frames from the movie. Question: [QUESTION] Answer it shortly and directly without repeating the question.'*

[13] https://huggingface.co/docs/transformers

[14] https://github.com/EvolvingLMMs-Lab/LongVA

[15] https://github.com/Vision-CAIR/LongVU

| Question | Answer |
|---|---|
| Setting-related | |
| Where do John and Emma have their conversation? | On the bus. |
| Which city is the backdrop? | Liverpool. |
| What decade does the movie take place in? | 1990s. |
| Character-related | |
| Who is the main character? | Josh. |
| What is the nationality of the soldier who becomes a prisoner? | German. |
| How is Mr. Jones described? | As stern and harsh. |
| Story-related | |
| Can you name a specific tactic Tommy uses against Tiny? | He steals Tiny's food at dinner and distracts her at a school-wide relay race. |
| What is the imminent threat to Greg's residence? | The house is in foreclosure and the bank is about to repossess it. |
| What triggers Carla to believe her husband Frank will surprise her for their anniversary? | A bouquet left on her doorstep. |
| Theme-related | |
| What is a major theme? | The theme revolves around reconnecting and the complexities of honesty between exes. |
| What main theme is explored through Sarah's experience? | The theme of romantic expectations versus present realities. |
| What is the main theme? | The processing of grief and reconciliation between siblings. |

Table S3: **Samples from SF20K-Test by question type.**

| **Template: "In [MOVIE TITLE], [QUESTION]"** |
|---|
| In 'Titanic', what happened to the Heart of the Ocean? |
| In 'The Godfather', how does Vito die? |
| In 'Harry Potter and the Chamber of Secrets', who is Dobby's master? |

Table S4: **Examples of data leakage queries for MovieQA.**

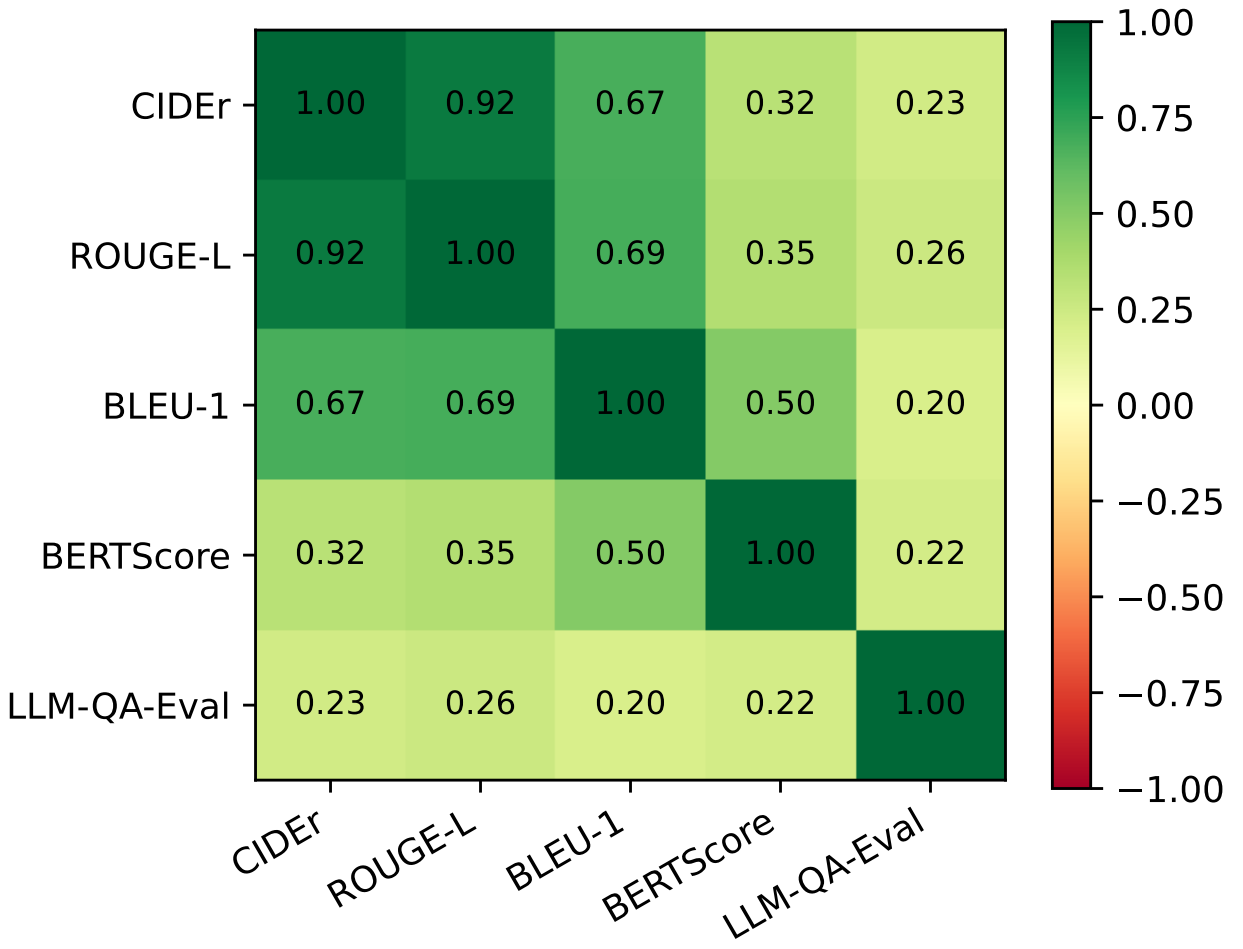

Fig. S3: **Inter-metric correlation** (Cohen's $\kappa$). N-gram metrics (*CIDEr*, *ROUGE-L*, *BLEU-1*) are highly correlated, while LLM-scoring (*LLM-QA-Eval*) stands apart. Text similarity (*BERTScore*) is only moderately correlated with BLEU-1.

We include audio information by adding the transcript in the prompt, in the following format: *'[START TIMESTAMP] - [END TIMESTAMP]: [SUBTITLE]'*. In case of modality ablation, we simply remove either the subtitles or the video frames from the inputs and adapt the query accordingly.

**Metrics.** For open-ended QA, we use an LLM to compare the free-form response from the model to the correct answer. In detail, we prompt GPT-4-nano (*'gpt-4.1-nano-2025-04-14'*) for the OpenAI API, and use the prompt provided by [54]. We get a yes/no value and an associated score. We then compute the average score and the overall accuracy.

## F Technical details on the temporal window study

In this section, we present additional details and results for the temporal window study (see Section 5.4). The goal of this experiment is to evaluate the performance of various methods on SF20K-Test at different temporal levels—shot, scene, and movie . We examine Qwen3-VL [5] at three model sizes: 2B, 4B, and 8B.

At the shot-level, we truncate the input prompt to include only information from a specific shot. For Qwen3-VL, that means that we only include subtitles and frames extracted from the shot. Therefore, we get a list of $n_{shots}$ predictions: we obtain a final prediction by using an LLM to pick the most probable answer, given the question as context.

The scene-level inference is similar to the shot-level one, the main difference being that we constrain the input information (subtitles and frames) to the scene-level, which arbitrarily corresponds to 10 shots in our case. We end up with a prediction list of length $n_{scenes}$. The movie-level inference is straightforward: it is the same as in Section 5.3.

## G Prompts

We provide the prompts for the data generation process: question generation prompt (see Table S5), and question classification prompt (see Table S6). As detailed in Section 4.1, those prompts were used with GPT-4 (*'gpt-4-turbo-2024-04-09'*) through the OpenAI API.

**Character identification**

*Identify the main characters, assign names, and recognize them in subsequent scenes.*

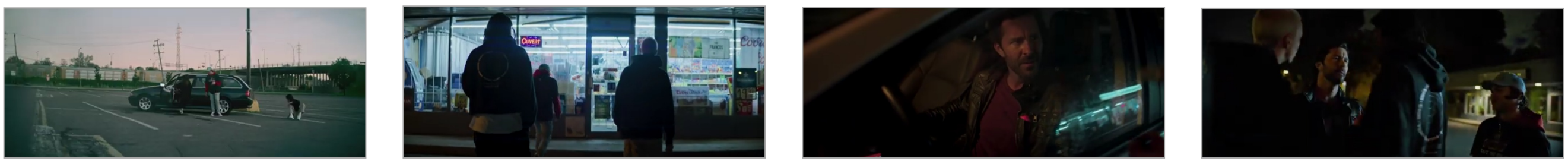

**Q** How many characters are there?
**✓ Answer** Four.
**✗ Prediction** Three.

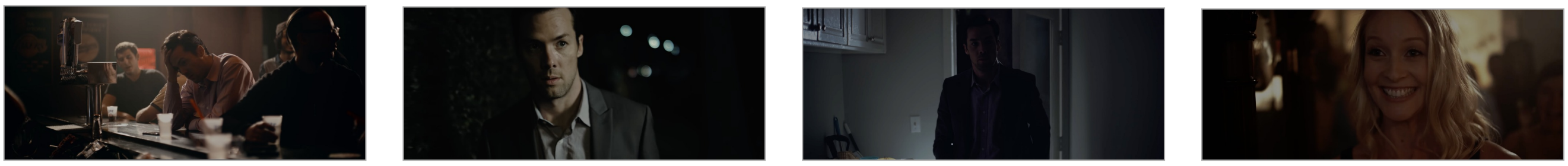

**Q** Whose birthday is it?
**✓ Answer** Vince.
**✗ Prediction** Jeff's.

**Long-range causal reasoning**

*Connect distant scenes to keep track of the overall story.*

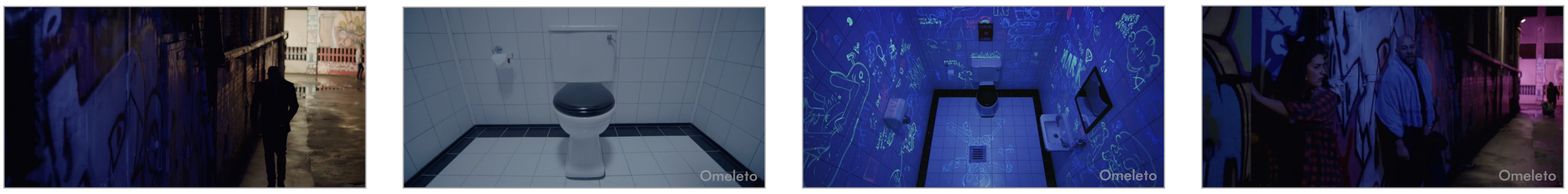

**Q** How do we know that the protagonist learned the lesson?
**✓ Answer** He tells a woman to keep the toilet clean.
**✗ Prediction** He finally follows the "Please be clean!" sign and leaves the restroom clean.

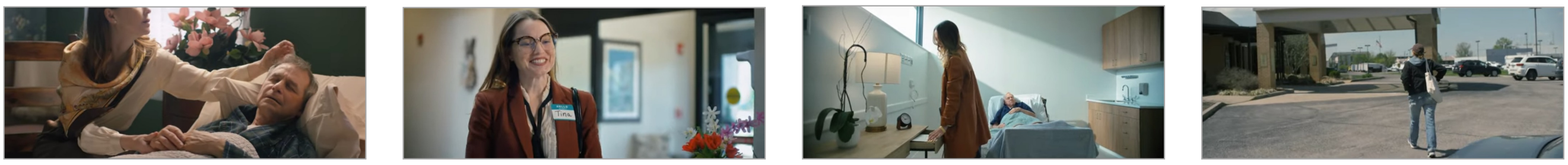

**Q** What does the protagonist try to steal from her third visit?
**✓ Answer** A cheque.
**✗ Prediction** A ring from her mother's hand.

**Visual story understanding**

*Perceive key visual elements that are required to understand the story.*

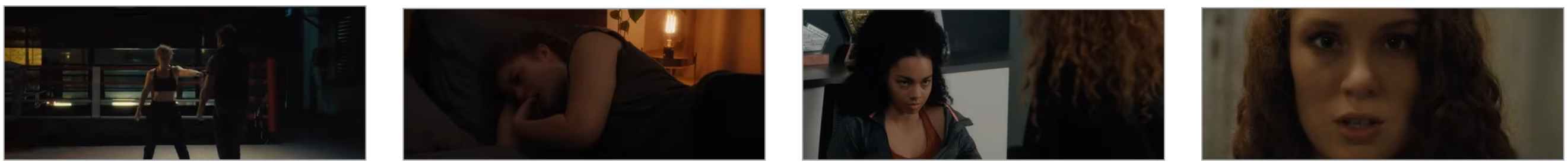

**Q** Which gesture is constantly repeating?
**✓ Answer** The gesture of disarming someone of their gun.
**✗ Prediction** Touching/rubbing her neck.

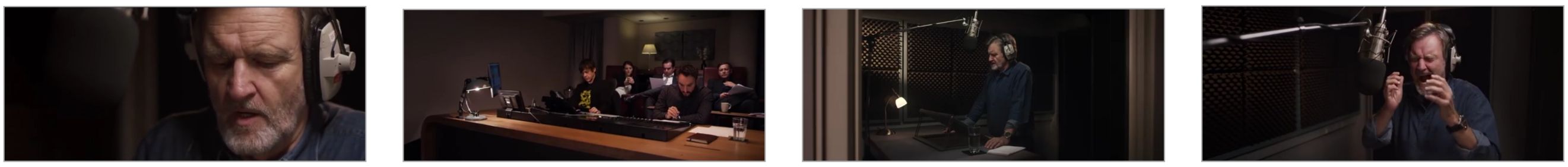

**Q** What is the action of the sound engineer when he understands what's happening with the final request?
**✓ Answer** He puts the volume to the maximum.
**✗ Prediction** He quietly hits record, keeping the mic rolling.

Fig. S4: **Qualitative failure cases.** We highlight three core challenges of story-level video understanding. For each example, we show a sequence of frames, the question, the ground-truth answer (✓), and the model's incorrect prediction (✗).

| Question Generation Prompt |
| --- |
| Movie title: {title}<br>Movie caption: {caption}<br>Movie synopsis: {synopsis}<br><br>Generate concise question-answer pairs based on the movie's synopsis, focusing specifically on the characters and plot, with the following guidelines:<br><br>Direct and Clear Questions: Craft each question to be straightforward, enabling someone who has watched the movie to answer without ambiguity.<br>Brief and Specific Answers: Provide answers directly derived from the synopsis, ensuring they are short and to the point. The aim is for these answers to be verifiable by watching the movie itself.<br>Handle Vague Details Carefully: If the synopsis mentions events without sufficient detail, avoid creating question-answer pairs about these specifics. Instead, focus on aspects that are clearly described.<br>Avoid Speculation: Do not infer or assume details not explicitly mentioned in the synopsis. Stick to the information provided.<br>Prevent Repetition: Ensure that each question is unique and does not overlap in content with others.<br><br>The goal is to create a set of questions and answers that accurately reflect the movie's content as described in the synopsis, suitable for testing comprehension through the movie viewing experience alone.<br>Please generate your response in JSON format. |

Table S5: **Question generation prompt.**

| Question Classification Prompt |
| --- |
| Classify the questions into one of the following category:<br><br>Setting-Related Question:<br>Question that focus on the location and time where the movie take place. It can be about either a geographical location (i.e. a specific city or country), a type of place (e.g. home, office, inside, outside), a historical period (i.e. a specific historical period), or a time of the day (e.g. morning, evening).<br><br>Character-Related Question:<br>Question that focus on the individuals portrayed in the movie. This can concern their personal information (e.g. age, gender, profession), theirs traits and motivations, and their relationships (e.g. siblings, friends). Note that we exclude character development, which is part of the narrative.<br><br>Story-Related Question:<br>Questions focus on the key sequence of events and interactions that make up the plot and narrative arc of the movie.<br><br>Theme-Related Question:<br>Questions focus on the underlying message or central idea that the movie explores and conveys to its audience.<br><br>Other:<br>Questions that do not fit into the above categories. These may include question about the director, the actors, the production, or other aspects of the movie that are not directly related to the story itself.<br><br>Question ID: {question_id}<br>Question: {question}<br>Answer: {answer}<br><br>Output format in JSON, as follows:<br>{{<br>'question_id1': 'question_type1',<br>}} |

Table S6: **Question classification prompt.**

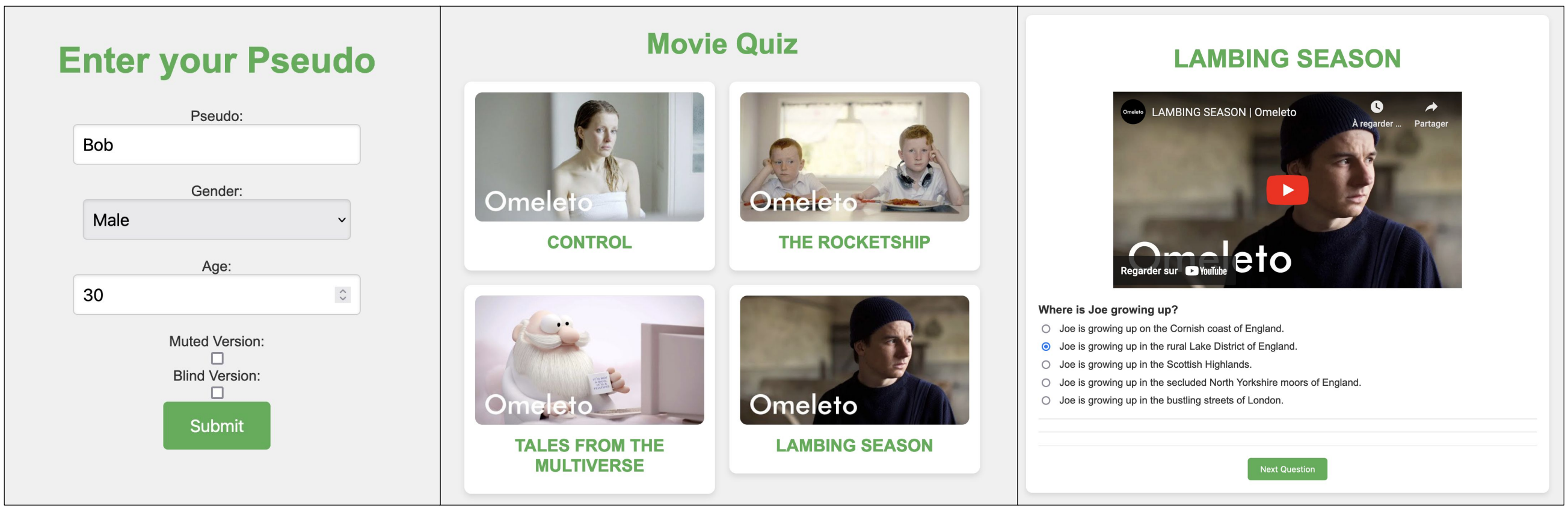

Fig. S5: **User study interface.** We ask participants to watch movies and answer the questions in order to assess human performance on our dataset.